\documentclass[journal]{IEEEtran}
\usepackage{amsmath,amsfonts}
\usepackage{algorithmic}
\usepackage{algorithm}
\usepackage{array}
\usepackage[caption=false,font=normalsize,labelfont=sf,textfont=sf]{subfig}
\usepackage{textcomp}
\usepackage{stfloats}
\usepackage{url}
\usepackage{verbatim}
\usepackage{graphicx}
\usepackage{cite}
\usepackage{hyperref}
\usepackage{soul, color}
\definecolor{hhhh}{RGB}{51,107,158}
\hypersetup{
    colorlinks=true,
    allcolors=hhhh,
}
\usepackage{caption}
\usepackage{ulem}
\usepackage{booktabs}
\usepackage{multirow}

\usepackage{enumitem}
\allowdisplaybreaks
\graphicspath{{imgs/}}
\DeclareGraphicsExtensions{.pdf, .png, .jpg}
\ifCLASSINFOpdf
\else
\fi

\begin{document}

\title{Neural Airport Ground Handling}

\author{Yaoxin Wu*, Jianan Zhou*, Yunwen Xia, Xianli Zhang, Zhiguang Cao, Jie Zhang
\thanks{*Yaoxin Wu and Jianan Zhou contributed equally to this paper.}
\thanks{This research was conducted in collaboration with Singapore Telecommunications Limited and supported by the Singapore Government through the Industry Alignment Fund ‐ Industry Collaboration Projects Grant. \emph{(Corresponding author: Zhiguang Cao.)}}
 \thanks{Yaoxin Wu is with the Department of Information Systems, Faculty of Industrial Engineering and Innovation Sciences, Eindhoven University of Technology, Netherlands. (E-mail: wyxacc@hotmail.com).}
 \thanks{Jianan Zhou, Yunwen Xia, Xianli Zhang and Jie Zhang are with the School of Computer Science and Engineering, Nanyang Technological University, Singapore (E-mails: jianan004@e.ntu.edu.sg, yunwen001@e.ntu.edu.sg, xianli001@e.ntu.edu.sg, zhangj@ntu.edu.sg).}
\thanks{Zhiguang Cao is with the School of Computing and Information Systems, Singapore Management University, Singapore. (E-mail: zhiguangcao@outlook.com).}
}

\markboth{IEEE Transactions on Intelligent Transportation Systems}%
{Shell \MakeLowercase{\textit{et al.}}: A Sample Article Using IEEEtran.cls for IEEE Journals}


\maketitle

\begin{abstract}

Airport ground handling (AGH) offers necessary operations to flights during their turnarounds and is of great importance to the efficiency of airport management and the economics of aviation. Such a problem involves the interplay among the operations that leads to NP-hard problems with complex constraints. Hence, existing methods for AGH are usually designed with massive domain knowledge but still fail to yield high-quality solutions efficiently. In this paper, we aim to enhance the solution quality and computation efficiency for solving AGH. Particularly, we first model AGH as a multiple-fleet vehicle routing problem (VRP) with miscellaneous constraints including precedence, time windows, and capacity. Then we propose a construction framework that decomposes AGH into sub-problems (i.e., VRPs) in fleets and present a neural method to construct the routing solutions to these sub-problems. In specific, we resort to deep learning and parameterize the construction heuristic policy with an attention-based neural network trained with reinforcement learning, which is shared across all sub-problems. Extensive experiments demonstrate that our method significantly outperforms classic meta-heuristics, construction heuristics and the specialized methods for AGH. Besides, we empirically verify that our neural method generalizes well to instances with large numbers of flights or varying parameters, and can be readily adapted to solve real-time AGH with stochastic flight arrivals. Our code is publicly available at: \url{https://github.com/RoyalSkye/AGH}.

\end{abstract}

\begin{IEEEkeywords}
Airport Ground Handling, Vehicle Routing Problem, Attention Model, Reinforcement Learning.
\end{IEEEkeywords}

\section{Introduction}
\label{intro}

\begin{figure}[!t]
    \centering
    \includegraphics[scale=0.125]{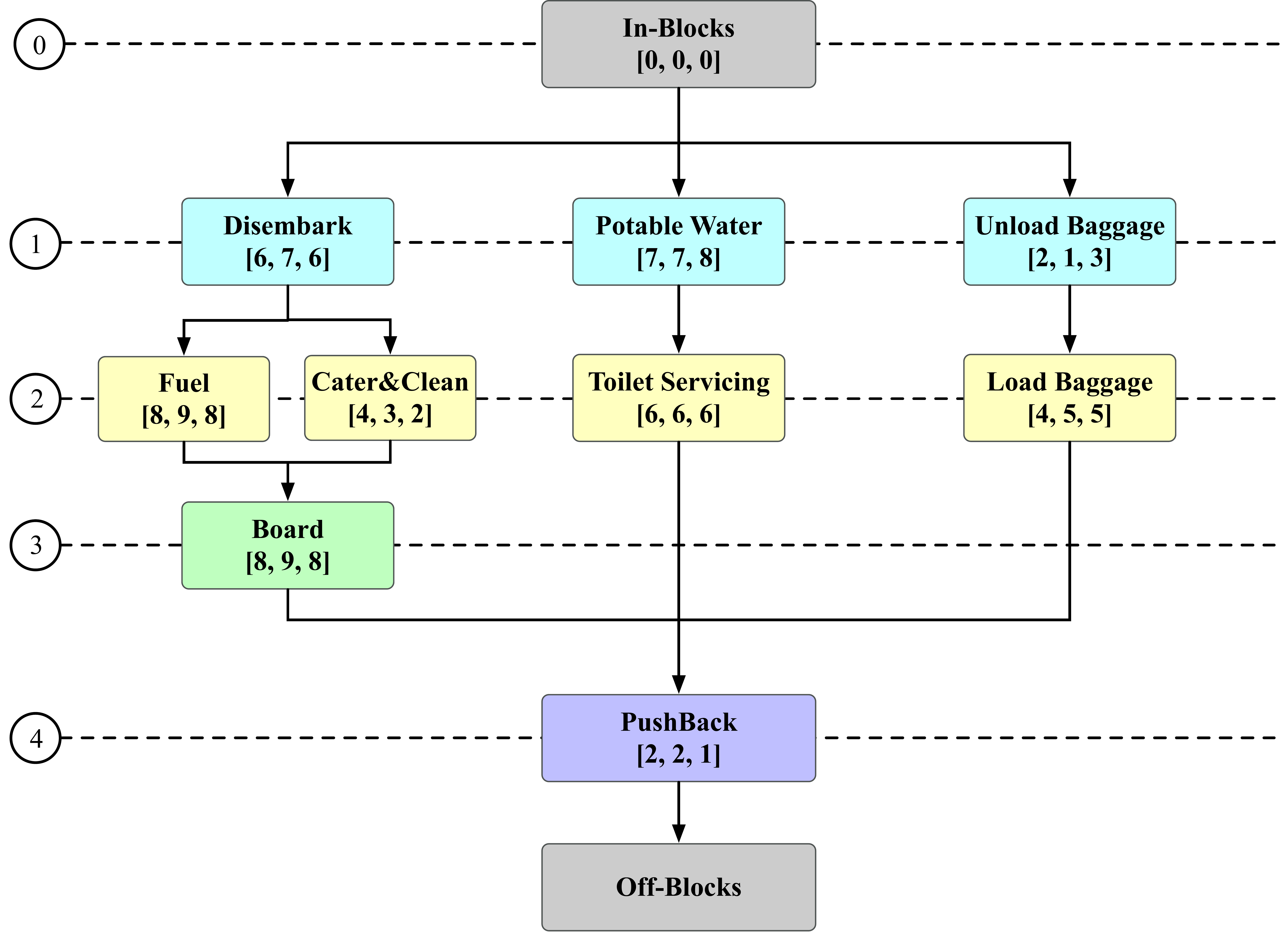}
    \caption{The exemplary operations in AGH and their precedence relation. The operations at the same level share the same priority, and a smaller level index means the corresponding operation has a higher priority, e.g., fueling, catering and cleaning have the same priority and precede boarding. Moreover, the numbers in the square brackets refers to the corresponding operation duration to three different flight types.}
    \label{fig_precedence}
    \vspace{-4mm}
\end{figure}

\IEEEPARstart{E}fficient turnaround operations for aircraft play an important role in alleviating the flight delays and relevant economic loss at an airport. In \emph{airport ground handling} (AGH), a variety of operations need to be scheduled to serve the flights according to their precedence relation (as shown in Fig. \ref{fig_precedence}), where each type of operation is always performed by a fleet of vehicles. Therefore, central to AGH is usually the vehicle routing problem (VRP). However, different from other conventional VRPs, the one in AGH is much harder given that, 1) apart from the intricate precedence constraint, each (type of) flight has its own requirement (e.g., demand, time windows) for each operation; 2) the problem scale is often large in practice especially for busy airports where a considerable number of flights need to be served at each time. Hence, solving the VRP in AGH is nontrivial and challenging (Note: hereafter, we use AGH to represent the VRP in AGH for convenience).

Considering the combinatorial nature of the problem, exact algorithms based on branch and bound (or its variants) have been exploited to solve AGH. They usually model the problem as mixed integer linear programming (MILP)~\cite{zhao2019bi,zhu2022cooperative}, and then solve it with mature solvers, e.g., CPLEX. However, such solvers are inefficient to handle the large and complex solution space, and usually consume prohibitively long computation time. On the other hand, as alternatives to balance the solution quality and computation time, traditional improvement heuristics or meta-heuristics also have been leveraged to solve AGH. In this line of works, a number of attempts design specific heuristics only for a single type of operation, e.g., towing, catering, fueling and baggage loading, respectively~\cite{du2014planning,ho2010solving,feng2021research,guo2020scheduling}. However, the problem could be much harder when multiple fleets for various operations are practically needed given their complex precedence relation. To handle this challenging setting, some works proposed specialized meta-heuristics to solve AGH with multiple operations~\cite{padron2016bi,liu2021scheduling}. Nevertheless, these methods depend on much trial-and-error and domain knowledge to design hand-crafted rules or operators, which may limit the solving performance and hinder their applications in reality.

Recently, an increasingly large number of deep models have been proposed to learn construction heuristics for solving VRP variants including traveling salesman problem (TSP) and capacitated vehicle routing problem (CVRP). Most of those deep models follow the encoder-decoder architecture and are trained with reinforcement learning (RL) algorithms~\cite{Bello2017WorkshopT, nazari2018reinforcement, kool2018attention, khalil2017learning, joshi2019efficient}. The learned construction policies infer solutions efficiently. More importantly, those end-to-end neural methods could produce desirable results which sometimes are even comparable to the highly specialized and optimized heuristics. Despite much success achieved, most of the neural methods are only able to tackle the classical VRP variants with much simple constraints, which will become less effective in coping with AGH.



In this paper, we propose a neural method that learns to construct solutions for AGH. Particularly, we first propose a general framework to decompose the global optimization of AGH into sub-problems for each fleet of vehicles based on the precedence of operations, so that a construction heuristic could be applied to solve them in sequence. However, those sub-problems may interplay each other due to the precedence relation (e.g., fueling goes before boarding) at each flight.
To address this issue, we present an attention-based policy network with masking schemes for constraints, where we explicitly encode time windows and operations in our neural network to capture the interplay. Moreover, we train the neural network with RL, which is shared across all the sub-problems. The resulting learnt policy is able to focus on long-term return and attain solutions of high-quality. Accordingly, our contributions are summarized as follows,
\begin{itemize}
  \item [1)] 
  We design a construction framework to decompose the AGH into sub-problems, which allows them to be sequentially solved with the proposed construction heuristic.
  \item [2)] 
  We parameterize the proposed construction heuristic with an attention-based policy network in conjunction with embeddings of time windows and operations. Further trained with RL and shared across sub-problems, it allows the learnt heuristic to automatically construct solutions for AGH without much hand-crafted rules. To our knowledge, this is one of the early neural methods to tackle VRPs with complex constraints.
  \item [3)] 
  We conduct extensive experiments with a realistic setting from the CHANGI Airport in Singapore. Results show that the proposed method efficiently produces high-quality solutions and outperforms the state-of-the-art conventional meta-heuristics. Particularly, our method can solve AGH with hundreds of flights in seconds, and is applicable to dynamic scenarios with stochastic arrivals.
\end{itemize}
The remainder of this paper is organized as follows. Section~\ref{related_work} reviews existing works related to AGH and neural methods for VRPs. Section~\ref{problem_statement} introduces AGH and its MILP formulation. Section \ref{methodology} elaborates the construction framework for AGH and the proposed neural method. Section \ref{exp} presents experimental results and analysis, and we conclude the paper in Section \ref{conclusion}.

\section{Related Works}
\label{related_work}
In this section, we first review the existing works pertaining to airport ground handling (AGH), and then provide a view of neural methods for general vehicle routing problems (VRPs).

\subsection{Airport Ground Handling}
Most existing works on AGH only tackle a specific (type of) operation, and model it as a vehicle routing problem (VRP) or related variants. Among them, Norin et al. formulated the scheduling of de-icing operation as a vehicle routing problem with time windows (VRPTW), and minimized both the tour length of de-icing vehicles and the total flight delay uisng the greedy randomized adaptive search procedure (GRASP) algorithm~\cite{norin2009integrating,norin2012scheduling}.
Du et al. modeled the scheduling of towing operation also as a VRPTW and solved it using the column generation method based on the MILP form of the problem~\cite{du2014planning}. 
Zhou et al. studied the scheduling of trailers considering both rolling windows and flight delays, which was solved using the genetic algorithm (GA)~\cite{zhou2018research}. Guo et al. considered scheduling vehicles for airport baggage transport, and solved it by leveraging the parameter or rule selection in GA~\cite{guo2020scheduling}. Han et al. described the scheduling of ferry vehicles as the form of MILP to minimize the number of used vehicles, which was solved using CPLEX~\cite{han2020optimal}. There were also a number of studies for scheduling of fueling operation that were solved with meta-heuristics like ant colony optimization (ACO) and iterated local search (ILS)~\cite{du2008aco,zampirolli2021simulated,feng2021research}.
On the other hand, only a few works investigated the global optimization for multiple (types of) operations in AGH. Padr{\'o}n et al. presented a decomposition method to schedule multiple interplayed operations based on constraint programming (CP) and large neighborhood search (LNS)~\cite{padron2016bi}. In particular, they first optimized the time windows for vehicle fleets using CP, then the decomposed VRPTW for each fleet were solved using LNS, which aimed to decrease the waiting time for operations and the total completion time for serving the flights. Moreover, they also ameliorated the method by solving the decomposed VRPTW in different orders, which impacted the performance due to the interplay relation among operations~\cite{padron2019improved}.  
A recent method leveraged non-dominated sorting genetic algorithm (NSGA-II) to solve the routing problems for multiple types of vehicles (fleets), with respective VRPTW modeled for each fleet~\cite{liu2021scheduling}. It aimed to minimize the total number of used vehicles as well as time cost of specified vehicles, where the presented method outperformed the multiobjective evolutionary algorithm based on decomposition (MOEAD) and the particle swarm optimization (PSO) algorithm. Other works less related to AGH in airside operation research and on-demand logistics, e.g., gate assignment, manpower scheduling, truck-drone coordinated VRPs, are introduced in~\cite{gok2020robust,gok2020simheuristic,andreatta2014efficiency,ng2018review,liu2022optimising,wang2022truck,liang2022survey}.

\subsection{Neural Methods for VRPs}
The advances of deep (reinforcement) learning have ignited considerable research on solving combinatorial optimization problems (COPs) with neural networks. The attempts in recent years have shown encouraging results in tackling various classic COPs~\cite{wu2021learning,zhang2020learning,lee2020fast,kwon2020pomo,gasse2019exact,chen2019learning}, where vehicle routing problems (VRPs) are receiving noticeably more attention than others~\cite{ma2022efficient,li2021hcvrp,ma2021learning,li2021heterogeneous,xin2021neurolkh,zhou2023learning}. Most of the deep models for VRPs follow the encoder-decoder architecture, where the encoder learns representations of problem instances, and the decoder constructs the solution (i.e., tours) sequentially. Among them, Vinyals et al. designed a RNN based sequence-to-sequence (seq2seq) neural architecture to solve TSP using supervised learning~\cite{vinyals2015pointer}. Bello et al. improved the training algorithm by changing it to reinforcement learning, which circumvents the needs of optimal solutions as the labels~\cite{Bello2017WorkshopT}. Moreover, they also introduced the masking scheme to guarantee feasible solutions, and attention glimpses to strengthen the network efficiency. 
Nazari et al. further transformed the encoder in the seq2seq model into the linear projection, and applied it to solve both TSP and CVRP~\cite{nazari2018reinforcement}. 

Rather than RNN, more subsequent works resorted to Transformer to rebuild seq2seq models using attention-based neural networks~\cite{vaswani2017attention}. Among them, Kool et al. adapted the attention modules (AM) in both encoder and decoder of Transformer to solve TSP, CVRP and other variants~\cite{kool2018attention}. Concurrently, Kaempfer et al. tailored the permutation invariant pooling layer in Transformer to seq2seq models and solved the multiple TSP \cite{kaempfer2018learning}. However, its performance was inferior to the AM in~\cite{kool2018attention}. On top of AM, Kwon et al. further enhanced the performance on both TSP and CVRP by breaking the symmetry that lies in the essentially same routes~\cite{kwon2020pomo}. Notably, many other variants of AM were also proposed recently given its desirable framework and convenience for implementation, e.g., the ones consider multiple relational attention~\cite{xu2021reinforcement} and multiple decoders~\cite{xin2021multi}. In addition, a number of endeavours focused on the graph representation for VRP. Khalil et al. described traditional greedy heuristics for TSP as MDP and trained a deep Q-Network (DQN) based policy network which embedded the states in episodes using Graph Neural Network (GNN) \cite{khalil2017learning,wu2020comprehensive}. Nowak et al. attempted to predict the solution to the graph based TSP using GNN and supervised learning~\cite{nowak2017note}. Joshi et al. proposed a graph convolutional network (GCN) with different decoding schemes to tackle TSP in both supervised or autoregressive fashion~\cite{joshi2020learning}. We would like to note that most of the above works fall into the category of neural \emph{construction} methods, which construct solutions (tours) for VRP sequentially (e.g., node by node). For another category of related works which learn \emph{improvement} heuristics or meta-heuristics, we refer the readers to surveys~\cite{mazyavkina2021reinforcement,karimi2022machine,bengio2021machine}


Although the aforementioned neural methods could automatically learn the rules to solve VRPs, most of them only handle problems with simple settings and constraints (e.g. TSP and CVRP). In contrast, this paper turns to solve a complex yet practical VRP with intricate constraints (i.e. AGH with multiple operations), and we propose an effective neural method to solve the problem by learning from data in an end-to-end fashion. To the best of our knowledge, it is a very early attempt to successfully solve the AGH with deep learning based method.

\section{Problem Statement}
\label{problem_statement}
In this section, we describe the AGH with multiple (types of) operations as a multiple-fleet vehicle routing problem with various constraints and formulate it as the form of mixed integer linear programming (MILP).

In practice, each (type of) operation for the flight is normally performed by a fleet of vehicles. Therefore, we model the AGH with multiple operations as a multiple-fleet VRP, additionally constrained by \emph{capacity}, \emph{time windows} and \emph{precedence}. Specifically, we represent the AGH on an undirected graph $\mathcal{G}=(\mathcal{N},\mathcal{E})$ with the node set $\mathcal{N}=\{0,1,\cdots,n, \dot{n}\}$ and edge set $\mathcal{E}=\{(i,j)|i,j\in \mathcal{N}; i\neq j\}$. Both node $0$ and $\dot{n}$ denote the depot but are used to differentiate the location which the vehicles start from and return to, so as to avoid temporal conflicts at the depot. We denote flights to be served as $\mathcal{N}^* =\mathcal{N} \setminus\{0, \dot{n}\}$ with demand $\delta_{i}^{f}$ for each $i\in \mathcal{N}^*$ and operation $f\in \mathcal{F}$ ($\mathcal{F}=\{1,\cdots,F\}$). Naturally, we set $\delta_0^f=\delta_{\dot{n}}^f=0$. For an operation $f\in \mathcal{F}$, each edge $(i,j)$ is assigned with a cost $c_{ij}^f$, with $c_{0 \dot{n}}^f=+\infty$ to avoid the meaningless travel between node $0$ and $\dot{n}$. In this paper, since we aim to minimize the global travel distance of all vehicles in all operations, we set $c_{ij}^1=\cdots=c_{ij}^F$, all of which equal to the distance between flight $i$ and $j$. For 
different operations, the precedence relation is represented as $f_1\prec f_2$ ($f_1, f_2\in\mathcal{F}$) if $f_1$ precedes $f_2$. We assume that each fleet\footnote{Note that fleets naturally share the same indices as operations since they are performed by the corresponding fleets. We also assume that the vehicles are homogeneous.} that corresponds to $f\in \mathcal{F}$ comprises vehicles $\mathcal{V}^f=\{1,\cdots,V^f\}$ with the capacity denoted by $Q^f$ and sufficiently large $V^f$. In addition, we denote by $d_i^f$ the service time for flight $i$ regarding operation $f$, which means how much time it takes to complete operation $f$ at flight $i$, with $d_0^f=d_{\dot{n}}^f=0$; we denote by $t_{ij}^f$ the travel time from flight $i$ to $j$ by the vehicle for operation $f$, with $t_{0\dot{n}}^f=0$; 
we denote by $[a_{i}^f, b_{i}^f]$ the time window to serve flight $i$, which means the start time of operation $f$ for serving flight $i$ should be between $a_{i}^f$ and $b_{i}^f$. Accordingly, the AGH is formulated as follows,

\begin{align}
  \label{eq1} \text{min.} \quad & \sum_{f\in \mathcal{F}}\sum_{v\in \mathcal{V}^f}\sum_{(i,j)\in \mathcal{E}} c_{ij}^fx_{ijv}^f\\
  \label{eq2} \text{s.t.} \quad & \sum_{i\in \mathcal{N}}\sum_{v\in \mathcal{V}^f}x_{ijv}^f=1, \forall j\in \mathcal{N}^*, f\in \mathcal{F} \\
  \label{eq3} & \sum_{i\in \mathcal{N}\setminus \{\dot{n}\}} \hspace{-2mm} x_{iuv}^f= \hspace{-2mm} \sum_{j\in \mathcal{N}\setminus\{0\}} \hspace{-3mm} x_{ujv}^f, \forall u\in \mathcal{N}^*, v\in \mathcal{V}^f, f\in \mathcal{F} \\
  \label{eq4} & \sum_{j\in \mathcal{N}^*}\sum_{v\in \mathcal{V}^f}x_{0jv}^f \leq V^f, \forall f\in \mathcal{F} \\
  \label{eq5} & \sum_{j\in \mathcal{N}^*}\sum_{v\in \mathcal{V}^f}x_{0jv}^f = \sum_{i\in \mathcal{N}^*}\sum_{v\in \mathcal{V}^f}x_{i\dot{n}v}^f, \forall f \in \mathcal{F} \\
  \label{conflict} &  \sum_{i\in \mathcal{N}\setminus\{0\}}\hspace{-1mm} \sum_{v\in \mathcal{V}^f}\hspace{-1mm} x_{i0v}^f=\hspace{-1mm} \sum_{j\in \mathcal{N}\setminus\{\dot{n}\}} \hspace{-1mm} \sum_{v\in \mathcal{V}^f}x_{\dot{n}jv}^f=0, \forall f\in \mathcal{F} \\    
  \label{capacity} & \sum_{i\in \mathcal{N}^*}\delta_i^f \sum_{j\in \mathcal{N}} x_{ijv}^f\leq Q^f, \forall v\in \mathcal{V}^f, f \in \mathcal{F} \\
  \label{eq8} & x_{ijv}^f(T_{iv}^f +d_i^f +t_{ij}^f - T_{jv}^f) \leq 0, \forall v\in \mathcal{V}^f, f \in \mathcal{F} \\
  \label{eq9} & a_i^f \leq T_{iv}^f \leq b_i^f, \forall v\in \mathcal{V}^f, f \in \mathcal{F} \\
  \label{eq10} & T_{iv}^{f_1} + d_i^{f_1} \leq T_{iv}^{f_2}, \forall f_1, f_2\in\mathcal{F}, f_1 \prec f_2\\
  \label{eq11} & x_{ijv}^f\in\{0,1\}, \forall (i,j)\in \mathcal{E}, v\in \mathcal{V}^f, f \in \mathcal{F} \\
  \label{eq12} & T_{iv}^f\geq 0, \forall i\in \mathcal{N}^*, v\in \mathcal{V}^f, f \in \mathcal{F}
\end{align}
where the decision variable $x_{ijv}^f$ determines whether a vehicle $v$ for operation $f$ serves flight $j$ after $i$, and $T_{iv}^f$ determines when to start serving flight $i$ by vehicle $v$ for operation $f$. Regarding the objective in Eq.~(\ref{eq1}), minimizing the global travel distance as the cost is considered in this paper. Regarding the constraints, Eq. (\ref{eq2}) ensures that each flight is only served by one vehicle for a given operation; Eq. (\ref{eq3}) ensures that a vehicle goes to serve a flight for an operation and will exactly leave the same fight; Eq. (\ref{eq4}) ensures that the number of vehicles to serve flights for an operation is not larger than the maximum number of vehicles in the fleet; Eq. (\ref{eq5}) ensures that the vehicles in a fleet which leave the depot will return to it; Eq. (\ref{conflict}) ensures that all routes start from and end at the depot; Eq. (\ref{capacity}) ensures that the demands fulfilled by a vehicle is not larger than its capacity; Eq. (\ref{eq8}) ensures the right temporal logic when the vehicle $v$ for operation $f$ serves two continuous flights $i$ and $j$. Specifically, the start time of its service at flight $j$ should not be earlier than the time point when the vehicle completes the service at flight $i$ and then arrives at flight $j$. Eq. (\ref{eq9}) ensures that the start time of operation $f$ at flight $i$ should be within the time window defined above. Eq. (\ref{eq8}) and (\ref{eq9}) together ensure that operations always start within the predefined time windows, which also obey the temporal sequence; Eq. (\ref{eq10}) ensures that the operations are performed to a flight following their precedence relation, where the exemplary precedence is depicted in Fig.~\ref{fig_precedence}. For example, the operation $f_2$ can only start at flight $i$ after that the operation $f_1$ is completed at the same flight, if $f_1$ precedes $f_2$. We further linearize Eq. (\ref{eq8}) as below,
\begin{equation}
\label{eq13} T_{iv}^f +d_i^f +t_{ij}^f - T_{jv}^f \leq C(1-x_{ijv}^f), \forall (i,j)\in E, f \in \mathcal{F},   
\end{equation}
where $C$ is a large constant (i.e. $10^6$), so that the studied AGH could be transformed into a mixed integer linear programming (MILP) problem.

Please note that this paper targets a more complex yet practical version of AGH in comparison to most existing literature, where multiple operations, hundreds of aircraft, and various relations (e.g. precedence and temporal relations) between operations and aircraft are considered. In the following section, we propose a learning based method to solve the AGH which we have modeled as a multiple-fleet VRP with diverse constraints. With the power of deep learning, our method can learn more effective heuristics and cost less domain knowledge and hand-engineering than the existing methods for AGH~\cite{padron2016bi,padron2019improved,liu2021scheduling}.  As we know, this is the first time that deep learning is applied to solve this practical VRP, rather than simple and standard VRPs in the literature such as TSP and CVRP.

\section{Methodology}
\label{methodology}
In this section, we introduce the proposed method to solve AGH. Firstly, we present a framework that is able to work elegantly with construction heuristics. It decomposes the global optimization of AGH into VRPs for each fleet based on the precedence relation, and sequentially constructs solutions (routes) to each resulting VRP which will also be used to update the time window of flights for the succeeding level of operations. Then we formulate the solution construction for each sub-problem as a Markov decision process (MDP), and parameterize the policy with an attention-based neural network, which is shared across all the sub-problems. Finally, we present the RL algorithm to train the policy network. The overview of our proposed method is illustrated in Fig. \ref{fig_overview}.

\subsection{Construction Framework for AGH}
\label{methodology_a}
\begin{figure*}[!t]
    \centering
    \includegraphics[scale=0.15]{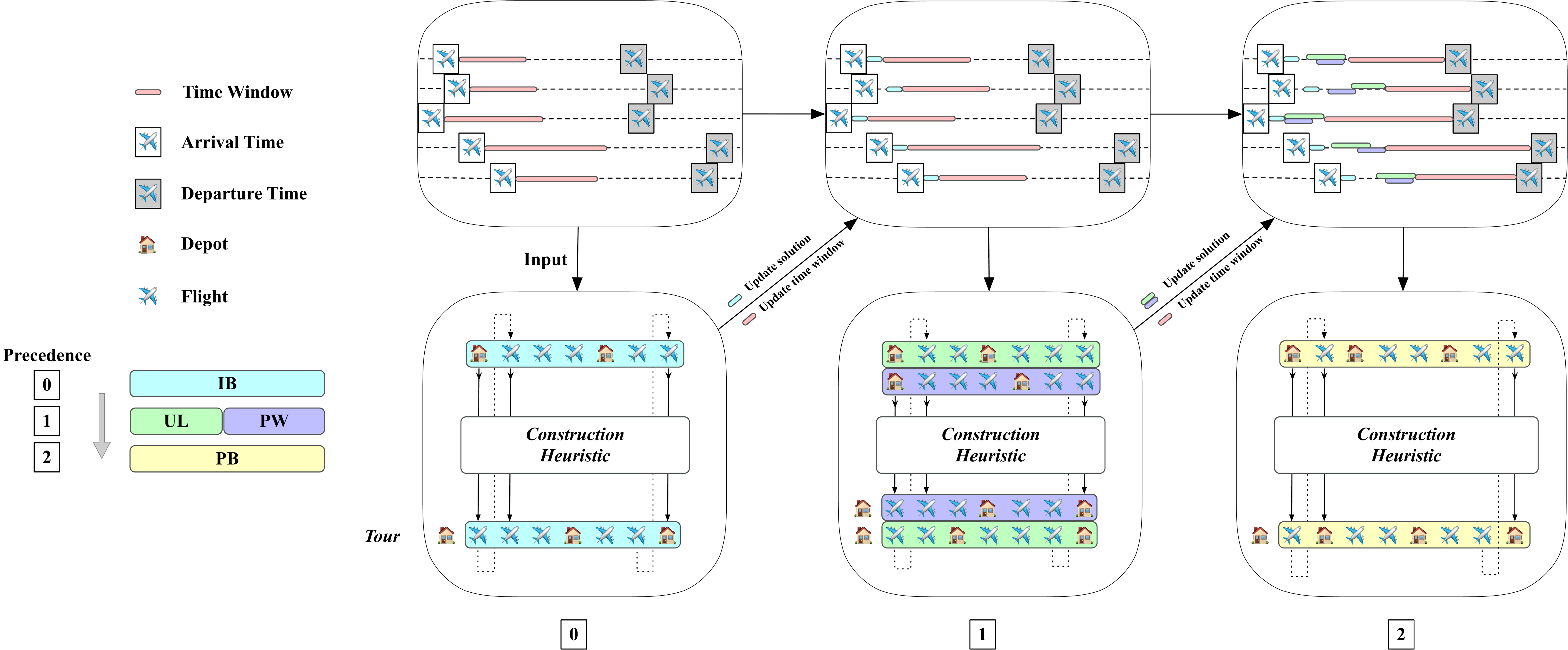}
    \caption{An illustration of the proposed construction framework for AGH. In this exemplary instance, five flights need to be served by four fleets of vehicles for corresponding operations with three precedence levels (priorities). Based on the precedence relation, we decompose this AGH into four sub-problems (or three groups). For each sub-problem, a local time window (red) of each flight is maintained, which constrains the start and completion of the current operation. The construction heuristic produces solutions for sub-problems, which are also used to update the time windows for VRPs in the next precedence level. 
    }
    \label{fig_overview}
\end{figure*}
The proposed framework for AGH is generic to construction heuristic such as Clarke and Wright's Savings (CWS), insertion based heuristics or learning based policies, which are used to solve the VRP for each fleet by selecting next flight to serve progressively. In specific, we decompose the AGH problem into sub-problems for each fleet. The sub-problems are essentially VRPs with capacity and time window constraints, which can be further grouped based on their precedence level (priorities). In doing so, we can solve the AGH by optimizing sub-problems following the precedence relation, which reduces the complexity and ensures the feasibility of solutions to the original problem. The \emph{decomposition strategy} and the \emph{overview of construction framework} is briefly summarized as follows (see also Fig. \ref{fig_overview}):
\begin{itemize}
  \item [1)] 
  Firstly, we decompose all fleets into several groups based on their precedence level. The fleets in each group should have the same precedence level, which means the time windows of the served flights for the current fleet will not be affected by other fleets in the same group. For example, the disembarkation, portable water supply and baggage unloading in Fig.~\ref{fig_precedence} are in the same group, which can be processed independently. In this way, the global AGH problem is decomposed into several groups, with each group comprising several sub-problems (fleets). 
  \item [2)] 
  Then we start with the group who has the highest precedence level. All sub-problems in the current group are solved by a solver or construction heuristic in the framework, sequentially or in parallel (i.e., the order doesn't matter since they can be processed independently). We solve the sub-problems in each group sequentially.
  \item [3)] 
  Based on the constructed solutions for the current group, we update the time windows of flights for fleets in the succeeding group whose precedence level is the most nearest to (and lower than) the current one. For example, the completion time of fueling, catering and cleaning will be used to update the time window of boarding in Fig.~\ref{fig_precedence}. In other words, the time windows for operations at flights are gradually determined by solving the sub-problems following the precedence relation.
  \item [4)]
  After solving all the groups, we obtain the global solution by simply collecting the solution to each sub-problem. It includes the derived routes for vehicles in each fleet, and the start time of operations can be easily computed based on the routes and operation duration. The global solution is feasible since all constraints are satisfied in the solving process. For example, the precedence constraint is satisfied since we solve sub-problems following the precedence level, and update the time windows accordingly; the capacity and time windows are satisfied when solving sub-problems by masking scheme as shown in Eqs. (\ref{eq:masking1}) and (\ref{eq:masking2}) in Section~\ref{methodology_c}.
\end{itemize}
Next, we formally introduce the details of the proposed construction framework for AGH.
For a set of operations $s(p)$ with precedence levels $p \in [0, ..., \mathcal{P}-1]$, we maintain a time window $tw_{j}^{p}=[a_{j}^{p}, b_{j}^{p}]$ for each flight $j$, where $a_{j}^{p}$ and $b_{j}^{p}$ represent the start and end of the time window, respectively. The operations in the precedence level $p$ must be launched and completed within the assigned time window.
Note that the time window of each flight is the same even for different operations if they have the same precedence level, such as fueling and catering in Fig.~\ref{fig_precedence}. Formally, the time window for operation $f$ at flight $j$ is constrained with the precedence level $p$ (i.e., $f \in s(p)$) and updated as follows,
\begin{gather}
\label{eq:tw_left_update}a_{j}^{p} = \begin{cases}
t_{a}^{j}, & p = 0 \\
\max_{f\in s(p-1)} T_{j}^{f}, & p \neq 0
\end{cases}
\end{gather}
\begin{gather}
\label{eq:tw_left_update2}b_{j}^{p} = t_{d}^{j} - \sum_{p\prec p'} {\max_{f \in s(p')} {d_{j}^{f}}},
\end{gather}
where $t_{a}^{j}$ and $t_{d}^{j}$ are the planned arrival and departure time of flight $j$; $T_{j}^{f}$ is the completion time of operation $f$ at flight $j$, which is calculated based on the constructed solutions for operations of the preceding precedence level; $d_{j}^{f}$ is the service duration of operation $f$ at flight $j$. Intuitively, the start of time window at a flight is initialized with the planned arrival time and updated based on the completion time of all operations with higher priorities. On the other hand, the end of time window is initialized with the departure time and progressively tighten based on the least reserved time required by the set of operations with lower priorities. The $\max$ operator in Eq. (\ref{eq:tw_left_update2}) results from the Cannikin Law\footnote{Also known as \emph{Wooden Bucket Theory}, where the capacity of a bucket is determined by the shortest stave.}, which means the least reserved time required for set $s(p')$ is determined by the operation with the longest duration, since it guarantees that all operations in $s(p')$ could be completed within $\max_{f \in s(p')} {d_{j}^{f}}$ in the optimal case.

As illustrated in Fig. \ref{fig_overview}, we deal with the precedence constraints and assign time windows to each flight following the aforementioned process in Eqs. (\ref{eq:tw_left_update}) and (\ref{eq:tw_left_update2}). The other constraints in VRPs for each fleet are handled by the construction heuristic. The derived solution for each fleet by the construction heuristic will also be used to update time windows of flights for operations in the succeeding precedence level. In this way, AGH is solved by constructing VRP solutions for fleets sequentially which will satisfy all constraints. Note that the operations with the same precedence level (e.g., fueling and catering in Fig. \ref{fig_precedence}) can be performed for the flight at the same time, and their solutions can be constructed in parallel with shared inputs (i.e., the same assigned time windows) to the construction heuristic.

\subsection{MDP Formulation}
\label{methodology_b}
According to the construction framework in Section~\ref{methodology_a}, the AGH problem $x$ can be decomposed into sub-problems $x^{f}$, where $f$ refers to an operation (or fleet). These sub-problems can be sequentially solved by any construction heuristic following the precedence relation as mentioned in Section~\ref{methodology_a}. Instead of classic construction heuristics, this paper aims to use DRL to learn construction policies to solve the sub-problems. To this end, for each sub-problem $x^{f}$, we formulate the solution construction as a Markov decision process (MDP), i.e., ($\mathcal{S}, \mathcal{N}, \mathcal{P}_a, \mathcal{R}_a$). Specifically, the \emph{agent} repeatedly selects the next flight to serve with the learnt policy, which results in a finite trajectory/episode. In particular, at the $t^{th}$ step, 
\begin{itemize}
    \item \emph{State:} $s_t^f\in S$ is represented by the static (i.e., the graph embedding) and dynamic information (i.e., the context embedding and previously selected flights). 
    \item \emph{Action:} $a_t^f \in \mathcal{N}$ is to select the next flight to serve w.r.t. the current operation according to the policy $\pi(a_t^f|s_t^f)$, where $\mathcal{N}$ is the node set $\{0, 1, \dots, n\}$ with $0$ denoting the depot and others for flight indices. 
    \item \emph{Transition:} $\mathcal{P}_a=P(s_{t+1}^f|s_t^f,a_t^f)$ refers to the transition from state $s_t^f$ to $s_{t+1}^f$, which is resulted from the action $a_t^f$. In our framework, the next state is realized deterministically w.r.t. the action, i.e., $\mathcal{P}_a=1$. 
    \item \emph{Reward:} After all flights are served within $T$ steps, the episode in MDP for operation $f$ ends, and a complete solution (tour) $a^f=(a_1^f, \dots, a_T^f)$ is attained. The total reward for the episode is represented as the negative length of the tour, i.e., $\mathcal{R}_a = -L(a^f)$.
\end{itemize}    

\subsection{Policy Parameterization}
\label{methodology_c}
\begin{figure*}[!t]
    \centering
    \includegraphics[scale=0.45]{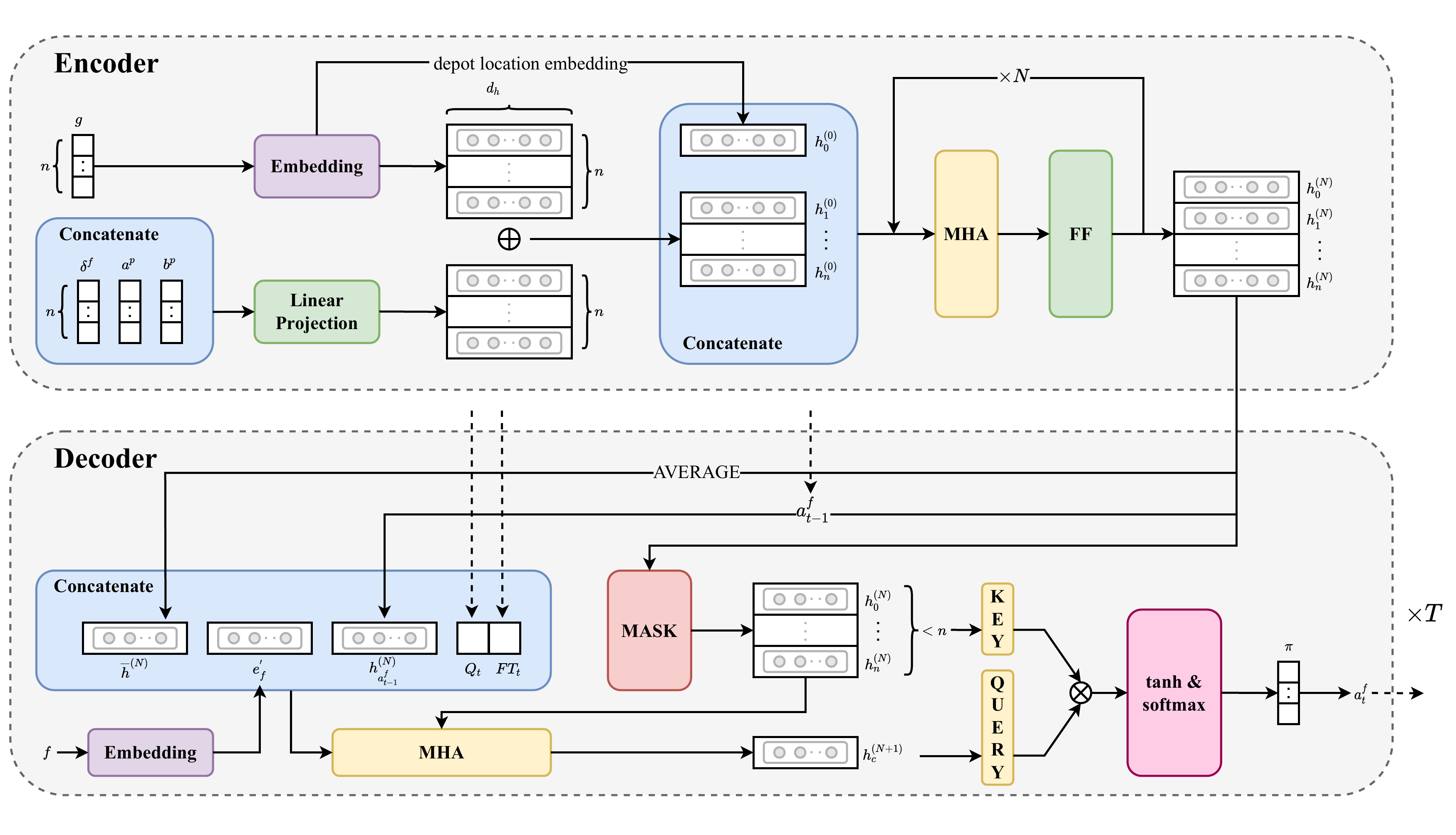}
    \caption{The architecture of the policy network to construct VRP solutions for respective operations in AGH.}
    \label{fig_model_structure}
\end{figure*}

We concretize the policy as an attention-based encoder-decoder model inspired by the attention model (AM) in~\cite{kool2018attention}, where the encoder learns representations of problem instances, and the decoder constructs solutions by learning which flight to select next for the current operation. Note that the learnt policy is shared across all the operations (sub-problems). Given a sub-problem $x^f$ for instance $x$, which is a VRP w.r.t. operation $f$, the probability of a solution $\pi_{\theta}(a^f|x^f)$ is parameterized by $\theta$ and factorized as follows, 
\begin{gather}
    \pi_{\theta}(a^f|x^f) = \prod_{t=1}^{T} \pi_{\theta}(a_{t}^f|x^f,a_{1:t-1}^f),
\end{gather}
where $a_{1:t-1}^f$ represents the current partial solution (route) that has been constructed before the $t^{th}$ iteration. At the $t^{th}$ iteration, we fix the previously constructed route $a_{1:t-1}^f$, and select an unvisited yet feasible flight (or depot) according to the policy $\pi_{\theta}$, which will be added into the current route. However, different from AM, our model handles much more complex VRP variants than the ones in~\cite{kool2018attention}. In particular, the architecture of our model is illustrated in Fig. \ref{fig_model_structure}, which primarily encompasses the encoder and decoder,

\subsubsection{Encoder} The encoder first computes the initial embedding $\textbf{h}_{j}^{(0)}$ for each node (i.e., flight and depot) with dimension $d_{h}=128$. On the one hand, we consider the location information of nodes. Different from the existing works where the coordinates of nodes in TSP or CVRP are randomly sampled according to a specific distribution, the flight nodes in AGH are located at some gates that are fixed in an airport. In other words, each flight is assigned to a specific gate for operations, and all possible locations of flights could be enumerated. Therefore, we leverage a learnable lookup table (i.e., a list of embeddings) to represent the location of each node. On the other hand, for each flight, we embed their demands and time windows by learning a linear projection. Then, the initial embeddings are attained by adding them together. Formally, the initial embeddings for depot and flight nodes are expressed as follows, 
\begin{gather}\label{eq:20}
    \textbf{h}_{j}^{(0)} = \begin{cases}
    \textbf{e}_0, & j = 0 \\
    \textbf{e}_{g_{j}} + \textbf{W}[\delta_{j}^{f}, a_{j}^{p}, b_{j}^{p}] + \textbf{b}, & j = 1, ..., n
    \end{cases}
\end{gather}
where $\textbf{e}_{g_{j}}$ is the embedding of the location $g_{j}$ for flight $j$; $\delta_{j}^{f}$ is the demand of flight $j$ for operation $f$; $a_j^p$ and $b_j^p$ are the start and end of time window of operation $f\in s(p)$ on flight $j$, respectively; $\textbf{W}\in \mathbb{R}^{d_{h}\times 3}$ and $\textbf{b}\in \mathbb{R}^{d_{h}\times 1}$ are trainable parameters; [$\cdot, \cdot$] is the horizontal concatenation operator. Note that since the time windows for the current operation are updated according to the solutions to the preceding operations, it is natural to leverage recurrent neural network (RNN) or its variants to specifically embed $a_{j}^{p}$ and $b_{j}^{p}$. However, it did not bring significant improvements over the linear projection w.r.t. the final global solutions. Hence, we stick to the usage of linear projection in Eq. (\ref{eq:20}) for better parallelism and higher computation efficiency (we discuss such a comparison in Section~\ref{exp_ablation}).

Given the yielded initial embeddings, the encoder updates them for $N$=3 times with the multi-head attention (MHA) ($M$=8 heads) layer and feed-forward (FF) layer, which further process their outputs using skip-connection and batch normalization (BN). Specifically, the update of embeddings at the the $l^{th}$ layer (Eq. (\ref{eq:21}) and~(\ref{eq:22})), and the forward pass of a MHA layer and FF layer (Eq. (\ref{eq:17})-(\ref{eq:19})) are expressed as follows, 
\begin{gather}\label{eq:21}
    \hat{\textbf{h}}_j^{(l)} = \text{BN}^{(l)} (\textbf{h}_{j}^{(l-1)} + \text{MHA}^{(l)}(\textbf{h}_{j}^{(l-1)};\textbf{h}_{0}^{(l-1)}, \dots, \textbf{h}_{n}^{(l-1)})),
\end{gather}
\begin{gather}\label{eq:22}
    \textbf{h}_{j}^{(l)} = \text{BN}^{(l)} (\hat{\textbf{h}}_j^{(l)} + \text{FF}^{(l)}(\hat{\textbf{h}}_j^{(l)})),
\end{gather}
\begin{gather}\label{eq:17}
    u_{jk}^m = \frac{(W_Q^m \textbf{h}_{j})^T (W_K^m \textbf{h}_{k})} {\sqrt{d_h/M}},
\end{gather}
\begin{gather}\label{eq:18}
    \hat{\textbf{h}}_j=\text{MHA}(\textbf{h}_{j};\textbf{h}_{0}, \dots, \textbf{h}_{n}) = \sum_{m=1}^{M} W_{O}^{m} [\sum_{k=0}^{n} \frac{e^{u_{jk}}}{\sum_{k'}e^{u_{jk'}}} W_{V}^{m} \textbf{h}_{k}],
\end{gather}
\begin{gather}\label{eq:19}
    \text{FF}(\hat{\textbf{h}}_j) = W_1^F \cdot \text{ReLu}(W_0^F\hat{\textbf{h}}_j + b_0^F) + b_1^F,
\end{gather}
where $l\in [1, \dots, N]$ is the layer index and $\textbf{h}_{j}^{(l)}$ is the embedding for flight $j$ at the $l^{th}$ layer; BN refers to the batch normalization; $(\textbf{h}_{0}, \dots, \textbf{h}_{n})$ are embeddings related to the depot and flight nodes; $W_Q^m$, $W_K^m$ and $W_V^m$ are trainable parameters to attain the query, key and value vectors in MHA; $W_{O}^{m}$ is the trainable parameters for linear projection in the $m^{th}$ head; $W_{0}^F$, $W_{1}^F$ and $ b_{0}^F$, $ b_{1}^F$ are trainable parameters for linear projection and bias in FF layer. The resulting advanced embeddings $\textbf{h}_{j}^{(N)}$ for nodes are then passed to the decoder to help decide the next flight to serve.

\subsubsection{Decoder} Given the embeddings from the encoder and the current partial solution $a_{1:t-1}^f$, the decoder sequentially selects a flight $a_t^f$ at time step $t$ to construct the VRP solution regarding each operation. To well represent the state $s_t^f$, we incorporate the context embedding $\textbf{h}_{c}^{(N)}$ into decoder to capture the dynamics at each step.
Formally, we define the context embedding at step $t\in[1,\dots, T]$ as follows,
\begin{gather}
    \textbf{h}_{c}^{(N)} = [\overline{\textbf{h}}^{(N)}, \textbf{e}_{f}^{'}, \textbf{h}_{a_{t-1}^f}^{(N)}, Q_{t}, FT_{t}],
\end{gather}
where $\overline{\textbf{h}}^{(N)}=\frac{1}{n+1}\sum_{j=0}^{n}\textbf{h}_j^{(N)}$ is the graph embedding, defined as the mean pooling of the advanced node embeddings from the encoder; $\textbf{e}_{f}^{'}$ is the fleet embedding used to distinguish the current operation $f$ from others (similar to the location embedding, we also use a learnable lookup table to represent it); $\textbf{h}_{a_{t-1}}^{(N)}$ is the embedding of the last (selected) node; $Q_{t}$ represents the remaining capacity of the current vehicle and $FT_{t}$ is the completion time on the last node by the current vehicle. Among them, while the graph embedding $\overline{\textbf{h}}^{(N)}$ and fleet embedding $\textbf{e}_{f}^{'}$ are fixed, others vary at each step during the solution construction for sub-problem $x^f$. We define $a_0^f = 0$ and start all tours of vehicles from the depot node, where each tour corresponds to a vehicle. We initialize $Q_t$ and $FT_t$ as $Q_{1}=1$ and $FT_{1}=0$ and update them according to Eqs. (\ref{eq:capacity_update}) and (\ref{eq:free_time_update}), respectively, 
\begin{gather}
    \label{eq:capacity_update} Q_{t+1} = \begin{cases}
    1, & j_t = 0\\
    Q_{t} - \delta_{j_t}^{f}, & j_t \neq 0
    \end{cases}
\end{gather}
\begin{gather}
    \label{eq:free_time_update} 
    FT_{t+1} = \begin{cases}
    0, & j_t = 0\\
    \max (FT_{t} + t_{j_{t-1} j_t}^f, a_{j_{t}}^{p}) + d_{j_{t}}^f, & j_t \neq 0
    \end{cases}
\end{gather}
where we define the node selected at current step as $j_t=a_t^f$, and the one at last step as $j_{t-1}=a_{t-1}^f$; $t_{j_{t-1}j_t}^f$ is the travel time from $j_{t-1}$ to $j_t$; and $\max (FT_{t} + t_{j_{t-1} j_t}^f, a_{j_{t}}^{p})$ refers to the start time of an operation on flight $j_t$, which is the larger value between the time when a vehicle arrives at flight $j_t$ and the start of the time window assigned to the flight. For example, if the latter is larger, a vehicle cannot serve the flight until $a_{j_{t}}^{p}$ (in Eq. (\ref{eq:tw_left_update})), i.e., the arrival time of the flight or the latest completion time of operations in the preceding precedence level. Otherwise, the former is larger and the selected flight is not to be served until the arrival of the current vehicle. Note that both $Q_t$ and $FT_t$ will be reinitialized if the vehicle visits the depot, which practically indicates that another vehicle starts from the depot to serve unvisited flights since the vehicles are homogeneous. However, this procedure is only used to showcase the tour construction. In practice, the vehicles in the same fleet could depart at the same time according to their respective routes to ensure that the earliest available flight could be served timely, while others could wait at the corresponding flights if needed. Moreover, our method also has favorable potential to handle heterogeneous vehicles as long as the relevant properties are properly embedded.

Given the context embedding $\textbf{h}_{c}^{(N)}$, we update it to $\textbf{h}_{c}^{(N+1)}$ through a MHA layer with message passing only from the flight nodes to the context node, which is similar to the \emph{glimpse} function in~\cite{Bello2017WorkshopT} as follows,
\begin{gather}\label{eq:update_context}
    \textbf{h}_c^{(N+1)} = \text{MHA}(\textbf{h}_{c}^{(N)};\textbf{h}_{0}^{(N)}, \dots, \textbf{h}_{n}^{(N)}),
\end{gather}
where we ignore FF layer, skip-connection and batch normalization (those used in encoder) for the sake of higher efficiency. On the other hand, the flight nodes should be masked during the update of the context embedding if they violate related constraints. For example, if flight $j$ has already been served in the previous step, then $\textbf{h}_{j}^{(N)}$ is not passed to the MHA layer in Eq. (\ref{eq:update_context}). 
Formally, the masking of the flight $m_t^j$ and depot $m_t^0$ at step $t$ are updated sequentially as follows,
\begin{gather}\label{eq:masking1}
    \footnotesize
    m_t^j = \begin{cases}
    {\rm FALSE}, & j \notin a_{1:t-1}^f \  {\rm and} \ \delta_{j}^{f} \leq Q_{t} \  {\rm and} \\ &, \max (FT_{t} + t_{j_{t-1} j}^f, a_{j}^{p}) + d_{j}^f \leq b_{j}^{p}\\
    {\rm TRUE}, & otherwise 
    \end{cases}
\end{gather}
\begin{gather}\label{eq:masking2}
    \footnotesize
    m_t^0 = \begin{cases}
    {\rm TRUE}, & j_{t-1}=0 \  {\rm and} \  \exists \, j \in \mathcal{N}-\{0\}, m_t^j={\rm FALSE}\\
    {\rm FALSE}, & otherwise 
    \end{cases}
\end{gather}
where \emph{TRUE} means the node is masked and thus cannot be selected at the current step. Specifically, the flight $j\in \mathcal{N}-\{0\}$ is masked if it has been already served by a vehicle regarding the current operation $f\in s(p)$, or its demand exceeds the remaining capacity of the vehicle, or the current vehicle cannot finish the service before the end of time window $b_{j}^{p}$. For the depot, we do not allow it to be immediately visited again if it has already been visited at last step while there are still unserved flights that can be visited without violating any constraint at the current step.

The score of selecting each action is realized through a single-head attention mechanism as described in Eq. (\ref{eq:action_value}). To satisfy constraints in AGH, the score of node $j$ needs to be masked as $v_t(j)=-\infty$ if $m_t^j$ = TRUE. Accordingly, the final probability distribution over candidate actions is calculated with the Softmax function as shown in Eq. (\ref{eq:prob_dist}).
\begin{gather}
    \label{eq:action_value}
    \footnotesize
    v_{t}(j) = \begin{cases}
    C \cdot \tanh (\frac{(\textbf{W}_{Q}\textbf{h}_{c}^{(N+1)})^T(\textbf{W}_{K}\textbf{h}_{j}^{(N)})}{\sqrt{d_h}}), & m_t^j=\text{FALSE} \\
    -\infty, & m_t^j=\text{TRUE}
    \end{cases}
\end{gather}
\begin{gather}
    \label{eq:prob_dist}
    \pi_{\theta}(a_t^f|x^f, a_{1:t-1}^f) = \frac{e^{v_{t}(a_t^f)}}{\sum_{j=0}^{n} e^{v_{t}(j)}}.
\end{gather}

Accordingly, our encoder-decoder structured neural network $\pi_\theta$ expresses a stochastic policy to generate a solution (tour) $a^f$ for the fleet (operation) $f$ given the instance $x^f$, that is, 
\begin{gather}
    \label{eq:chain_rule}
    \pi_{\theta}(a^f|x^f) = \prod_{t=1}^{T}\pi_{\theta}(a_t^f|x^f, a_{1:t-1}^f),
\end{gather}
where $T$ means the number of steps used to construct a feasible solution (tour) which serves all aircraft.

\subsection{Policy Training}
Given the stochastic policy parameterized by the neural network in Section \ref{methodology_c}, with its solution expressed in Eq. (\ref{eq:chain_rule}), we train the policy network by exploiting the REINFORCE algorithm~\cite{williams1992simple} with a rollout baseline~\cite{sutton2018reinforcement}. The pseudocode of training procedure is displayed in Algorithm \ref{alg:reinforce}, where $L_i^f$ and $\mathcal{B}_i^f$ mean the cost (length) of the constructed route (tour) for fleet $f$ in the $i$-th instance by the current trained model and the best saved one, respectively. Correspondingly, $\{\{L_i^f\}_{i=1}^B\}_{f=1}^F$ and $\{\{\mathcal{B}_i^f\}_{i=1}^B\}_{f=1}^F$ mean the list of costs in all operations and instances by the two models, respectively. After each epoch, we use the best checkpoint (i.e. the best saved model so far) to greedily solve instances in the validation set $X^{val}$,\footnote{We randomly generate
1000 instances after each epoch with the same size and random seed as those used in training to evaluate the current model.} and the current model is also used to solve the set. We substitute the current one for the best checkpoint if it can achieve significantly better performance according to a paired t-test~\cite{kool2018attention}, as shown in Line 14$\sim$17 in Algorithm \ref{alg:reinforce}. For each fleet w.r.t. operation $f$, the loss is defined as the expected solution cost as follows,
\begin{gather}\label{eq:train}
    \mathcal{L}(\theta|x^f) = \mathbb{E}_{\pi_{\theta}(a^f|x^f)}[L(a^f)],
\end{gather}
where $L(a^f)$ means a function to compute the length of the constructed route (tour) $a^f$. Note that since we explicitly embed the vehicle and operation information, it is expected that the learned policy is robust and reliable enough to also tackle VRPs for the heterogeneous fleets (i.e. operations) in AGH. To this end, we perform the gradient back-propagation, with the solution to each fleet $a^f$ separately constructed by the same policy network in Section \ref{methodology_c}. Formally, we optimize the policy network for solving AGH with the following update scheme,
\begin{algorithm}[!t]
  \caption{Policy Optimization with REINFORCE}
  \label{alg:reinforce}
	\begin{algorithmic}[1]
		\STATE \textbf{Input}: number of epochs $E$; number of iterations per epoch $I$; number of operations $F$; batch size $B$; significance $\alpha$.
		\STATE Init $\theta$, $\theta^{*} \gets \theta$
		\FOR{$epoch = 1, ..., E$}
		    \FOR{$iter = 1, ..., I$}
		        \STATE generate random instances $\{x_i\}_{i=1}^B$
		        \STATE $\{\{\mathcal{B}_i^f\}_{i=1}^B\}_{f=1}^F \gets$  Greedy($\{x_i\}_{i=1}^B, \theta^{*}$)
		        \FOR{$f = 1, ..., F$}
		            \STATE $\{L_i^f\}_{i=1}^B \gets$  Sample($\{x_i^f\}_{i=1}^B, \theta$)
		            \STATE update time windows as Eqs. (\ref{eq:tw_left_update}) and (\ref{eq:tw_left_update2})
		        \ENDFOR
		        \STATE obtain gradient $\frac{1}{B}\nabla\mathcal{L}$ using Eq. (\ref{gradient})
		        \STATE $\theta \gets \text{Adam}(\theta, \frac{1}{B}\nabla\mathcal{L})$
		    \ENDFOR
		    \STATE $\{\{L_i^f\}_{i=1}^{|X^{val}|}\}_{f=1}^F \gets$ Greedy($X^{val},\theta$)
		    \STATE $\{\{\mathcal{B}_i^f\}_{i=1}^{|X^{val}|}\}_{f=1}^F \gets$ Greedy($X^{val},\theta^{*}$)
		    \IF{t-test($\sum_{f=1}^{F}\{L_i^f\}_{i=1}^{|X^{val}|}, \sum_{f=1}^{F}\{\mathcal{B}_i^f\}_{i=1}^{|X^{val}|}) < \alpha$}
		        \STATE ${\theta}^{*} \gets \theta$
		        \STATE generate a new validation set $X^{val}$
		    \ENDIF
		\ENDFOR
	\end{algorithmic}
\end{algorithm}
\begin{equation}\label{gradient}
\begin{aligned}
\nabla\mathcal{L}(\theta|x) &= \sum_{x\in \{x_i\}_{i=1}^B}\frac{1}{|\mathcal{F}|} \sum_{f \in \mathcal{F}} [(L(a^f)\\
&\quad\quad\quad\quad-\mathcal{B}(x^f)) \nabla \log \pi_{\theta}(a^f | x^f)],\\
\theta &\gets \text{Adam}(\theta,\frac{1}{B}\nabla\mathcal{L}),
\end{aligned}
\end{equation}
where $\mathcal{B}(x^f)$ represents the length of the solution to operation $f$ constructed by the baseline (i.e. the best checkpoint) in a greedy manner. We use the above gradient to update the parameters of the policy network, as shown in Line 5$\sim$12 in Algorithm \ref{alg:reinforce} Intuitively, the above training diagram does not explicitly handle the interplay between solutions to different sub-problems. Alternatively, we also try to update the policy network with the global solution, which is constructed by using the network to schedule all operations in one go (rather than separate computations for each sub-problem). However, we observe that this alternative paradigm yields similar performance, which in turn increases additional training complexity with a long episode to construct the solution. We compare and analyze those different training paradigms in Section \ref{exp_ablation}.

In summary, we propose a deep learning based construction framework to solve AGH, with constraints properly handled for solution feasibility, the policy network elegantly structured for informative representation and a novel training algorithm to learn effective policies. Compared to use cases of RL for solving simplified VRP (e.g. TSP or CVRP) in the literature, this is the first time that RL is designed and applied to the complex yet practical VRP in AGH. Extensive experiments in the next section will show that the proposed method achieves significantly better performance than classic methods.

\section{Experiments and Analysis}
\label{exp}

In this section, we evaluate our neural method on AGH instances with 20, 50, 100 flights of 3 types, and 10 types of operations. We compare our method with classic meta-heuristics, construction heuristics and recent methods specifically designed for AGH to verify the superiority. Besides, we evaluate the generalization of the learned policy on instances with larger problem sizes and different parameters, and conduct ablation studies to assess the effectiveness of key designs in the proposed method. Moreover, we also adapt our method to real-time setting with stochastic arrival of flights to verify its capability to handle dynamics.

\subsection{Experimental Settings}
\label{exp_settings}
\noindent\textbf{Instance Generation.} We take the Changi Airport in Singapore as a use case. It involves 3 terminals and 91 gates, and we assume the depot node is placed at a specified location. To conduct the experiment, we load the airport map\footnote{\url{https://www.changiairport.com/en/maps.html}} into SUMO~\cite{SUMO2018}, where the distances among locations could be calculated automatically. On top of that, we generate instances with 20, 50 and 100 flights, which are referred as AGH20, AGH50 and AGH100. In those instances, we assign gates randomly to each flight. We consider 3 types of flights (each flight is assigned to a type randomly) and sample their arrival time according to the statistics\footnote{\url{https://www.changiairport.com/en/flights.html}} of real scenarios in Changi airport. Based on the arrival time, the departure time is computed by adding the duration of a turnaround (i.e., 30 minutes, 34 minutes and 33 minutes to the 3 flight types, respectively), which are sufficiently long in general. Regarding operations, we consider the 10 operations, their service duration and precedence relations in Fig.~\ref{fig_precedence} for all instances. Pertaining to each operation, the demand of flights are uniformly sampled from $[1, \dots, 9]$. We set the capacity of vehicles in each fleet to 30, 40 and 50 in AGH20, AGH50 and AGH100, respectively. Note that we normalize the demand by the capacity of vehicles before feeding them to neural networks, and hence the initial capacity is 1 as shown in Eq.~(\ref{eq:capacity_update}). We set vehicle speeds following \cite{de2010ground}. For each problem size (i.e. AGH20, AGH50, AGH100), we train the policy network with 100 epochs and generate 12800 instances in each epoch. We randomly generate 1000 instances with the same size and random seed after each epoch as the validation set to evaluate the current model, in order to keep the best-so-far model. For testing, we generate 1000 instances of AGH20, AGH50 and AGH100 with a random seed that is different from the one for training. Similarly, we generate 1000 instances of AGH200 and AGH300 for evaluating the generalization across problem sizes. In addition, we also test our method to solve AGH instances with different parameters and real-time settings, with the instance generation introduced in the corresponding section (i.e. Section \ref{exp_generalization} and \ref{stochastic_agh}).

\noindent\textbf{Baselines.} We compare the learned policy with representative meta-heuristics, construction heuristics and recent specialized methods for AGH. All baselines with their implementation details are introduced as follows.
\begin{itemize}
\item CPLEX (v20.1), a mature and highly optimized commercial solver for mixed integer linear programming (MILP)~\cite{cplex2009v12}. 
\item Insertions, a type of construction heuristics to solve VRPs. We compare with random, nearest, farthest insertion and nearest neighbor~\cite{braysy2004evolutionary}.
\item Clarke and Wright Savings (CWS), a construction heuristic to solve VRPs~\cite{solomon1987algorithms}. The first flight to be served is selected according to the start of time windows, and the route is constructed progressively according to the distance savings and temporal relations between every two flights.
\item Simulated Annealing (SA), a meta-heuristic that mimics the physical annealing process~\cite{franzin2019revisiting}. We use the nearest neighbor heuristic to find the initial solution, and equip it with \emph{swap} operator to explore the solution space (since it is better than \emph{2-opt} for AGH according to our preliminary results). We use Metropolis criterion to accept solutions. Concretely, we run SA for each fleet independently. We set the size of the neighborhood (defined by the swap operator) to 500, the maximum iteration number to 100, the initial temperature to 200, the cooling factor to 0.9, and the time limit to 30m.
\item Large Neighborhood Search (LNS), a meta-heuristic that generalizes neighborhood search for optimization by iteratively refining an incumbent solution with local search~\cite{pisinger2019large}.
Instead of explicitly defining a neighborhood function (e.g., k-opt operation), LNS defines the neighborhood implicitly through a pair of destroy and repair operators.
For each instance, we obtain an initial solution with a simple heuristic based on the nearest neighbor heuristic. Following the LNS framework in~\cite{song2020general}, the destroy operator iteratively selects decision variables in MILP to be reoptimized, and the repair operator (i.e., CPLEX) attains a feasible solution in the reoptimization. The feasible solution replaces the incumbent one to be further improved afterwards if it has a better objective value. Concretely, in each LNS iteration, we randomly select 50 percent of decision variables to be destroyed (and reoptimized) while fixing others. The time limit of each LNS iteration is set to 20s, 1m and 2m for AGH20, AGH50 and AGH100, respectively.
\item LNS\_SA, a natural combination of LNS and SA where SA is used as the acceptance criterion in LNS framework. Specifically, instead of simply greedily accepting new feasible solutions, we use Metropolis criterion to accept solutions as used in the SA. The initial temperature is 200, and is cooled by 0.95 every 10 LNS iterations. The other settings (e.g., the time limit of each LNS iteration) remain the same as the ones described above.
\item Genetic Algorithm (GA), a recent method specifically designed for AGH~\cite{liu2021scheduling}. While the original work only considers one type of aircraft, we adapt it to our setting with heterogeneous aircraft. The solution (chromosome) of each fleet is represented as an array, where the first half represents the vehicle index required by each flight, and the last half represents the order in which each vehicle serves the flight. The selection, crossover and mutation operators are the tournament selection, simulated binary crossover and polynomial mutation, respectively. The hyperparameters follow the ones used in \cite{liu2021scheduling}.
\item CP\_LNS, which tackles similar AGH to ours by first decomposing it into sub-problems for each fleet according to temporal constraints, and then applying LNS with constraint programming (CP) to solve them~\cite{padron2016bi,padron2019improved}. We adapt it to solve exactly the same AGH to ours.
The neighborhood structure is defined by two operators: the Random Pivot operator (RPOP), which removes and reinserts individual customers, and the Small Routing (SMART), which relies on arc exchanges. The time limit of each LNS iteration is kept the same as the vanilla LNS described above.
\end{itemize}

\begin{table}[!t]
  \newcommand{\tabincell}[2]{\begin{tabular}{@{}#1@{}}#2\end{tabular}}
  \renewcommand{\arraystretch}{1.8}
  \renewcommand\tabcolsep{3.5pt}
  \caption{Performance regarding ablation studies.}
  \label{exp_table_3}
  \centering
  \footnotesize
  \begin{tabular}{l|rrr|rrr}
  \toprule
   \multirow{2}*{Method} &
   \multicolumn{3}{c|}{AGH100 (Greedy)} &
   \multicolumn{3}{c}{AGH100 (Sampling)}\\
    & Obj. & Gap & Time & Obj. & Gap & Time\\
   \hline
   $\mathcal{L}_{G}$ & 576733.94 & 139.53\% & 4.09s & 531477.94 & 136.92\% & 5.35s \\
   $\mathcal{L}_{M_{G}}$ & 363241.50 & 50.73\% & 4.10s & 316558.09 & 41.02\% & 5.28s \\
   $\mathcal{L}_{M_{F}}$ & 299984.50 & 24.47\% & 4.10s & 263335.75 & 17.29\% & 5.27s \\
   \hline
   w/o TW & 338958.66 & 40.72\% & 4.12s & 301930.75 & 34.53\% & 5.24s \\
   w. LSTM & 245940.63 & 2.01\% & 4.22s & 227446.69 & 1.27\% & 5.34s\\ 
   \hline\hline
   Ours & \textbf{242930.86} & \textbf{0.75\%} & 4.08s & \textbf{225617.61} & \textbf{0.45\%} & 5.13s \\
  \bottomrule
  \end{tabular}
\end{table}

We report the average performance in terms of objective values and relative gaps\footnote{In specific, let $\bar{x}$ and $\bar{x}^*$ be the solution found by the current method and the best solution found among all methods, respectively. For each instance, we define the gap of $\bar{x}$ to the best found solution as: 
$|c(\bar{x})-c(\bar{x}^*)|/c(\bar{x}^*)$, where $c(\cdot)$ is the cost of a solution. We report the average gap over all instances in the testing set throughout the experiment.}
of all methods on the testing set, where the gaps are calculated against the best solution found among all the methods. For our method, we use two decoding strategies, 1) \textit{Greedy}, which always selects the flight with the highest probability (based on the output of policy network) at each step; 2) \textit{Sampling}, which samples 1000 solutions in parallel based on the output probability distribution of policy network, and selects the best one as the final solution. 
We also report the average time needed to solve a single instance, the setting of which may vary with different methods. Note that we implement (run) all methods on CPU except that our neural method is on GPU. In specific, we run those experiments on a server equipped with a single GeForce RTX-2080Ti GPU card and Intel i9-10940X CPU @ 3.30GHz.

\noindent\textbf{Hyperparameters.} Regarding CPLEX and meta-heuristics, we set time limit as 30 minutes to solve instances of AGH20, AGH50 and AGH100. In addition, for GA, we set the population size to 100, crossover probability to 0.7 and mutation probability to 0.3. 
For SA, we explore 500 candidate solutions in the neighborhood of current solution at each iteration of the local search. The initial temperature is set to 200 with decay rate of 0.9. For LNS, we set the degree of destruction to $0.5$, which means we reoptimize a sub-MILP with half unfixed decision variables (while the other half is fixed) in the original MILP at each iteration. Moreover, we use CPLEX as the repair operator in LNS. For each LNS iteration, the time limit for AGH20, AGH50 and AGH100 is set to 20 seconds, 1 minute and 2 minutes, respectively. For LNS\_SA, the initial temperature is set to 200 with decay rate of 0.95 for every 10 LNS iterations. For our neural method, we train the policy network for 100 epochs with instances generated on the fly. In each epoch, 200 batches of 64 instances are processed for training, and 1000 instances are evaluated at the end of each epoch. In the neural structure, we use $N=3$ layers in the encoder and $M=8$ heads in each multi-head attention layer. The dimension of all embeddings is set to 128. We use Adam optimizer~\cite{kingma2015adam} to update the parameters of neural network with a constant learning rate $10^{-4}$. The significance $\alpha$ in t-test is set to $5\%$.

\begin{figure}[!t]
    \centering
    \includegraphics[width=1.\linewidth,height=5.5cm]{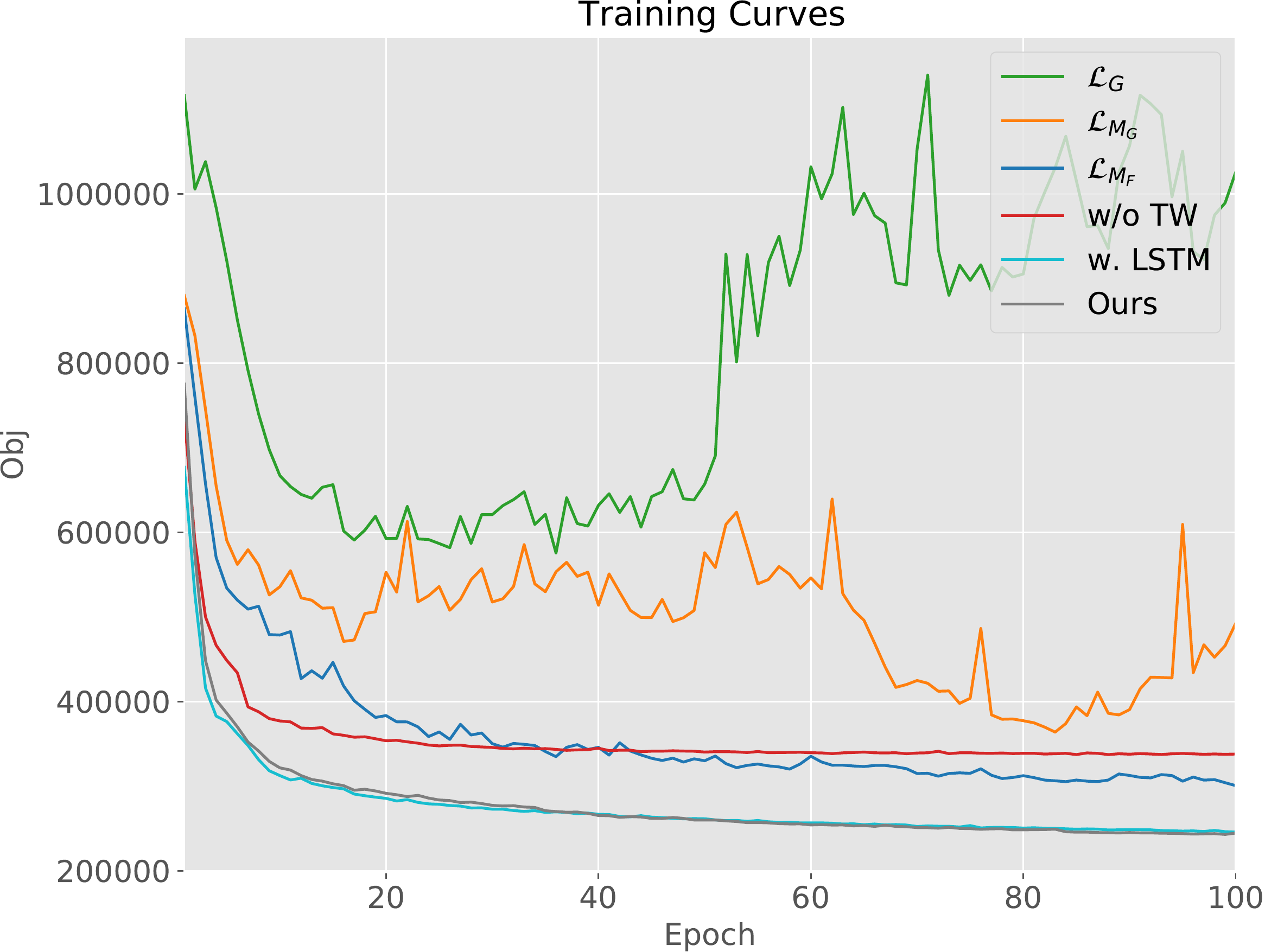}
    \caption{Training curves regarding ablation studies.}
    \label{fig_ablation}
    \vspace{-3mm}
\end{figure}

\begin{table*}[!t]
  \newcommand{\tabincell}[2]{\begin{tabular}{@{}#1@{}}#2\end{tabular}}
  \renewcommand{\arraystretch}{1.5}
  \renewcommand\tabcolsep{7.0pt}
  \caption{Comparison with baselines.}
  \label{exp_table_1}
  \centering
  \footnotesize
  \begin{tabular}{l|rrr|rrr|rrr}
  \toprule
   \multirow{2}*{Method} & \multicolumn{3}{c|}{AGH20} & \multicolumn{3}{c|}{AGH50} & \multicolumn{3}{c}{AGH100}\\
    & Obj. & Gap & Time & Obj. & Gap & Time & Obj. & Gap & Time \\
   \hline
   CPLEX & \textbf{80151.01} & \textbf{0.00\%} & 30m & 300801.58 & 104.09\% & 30m & 795687.22 & 252.23\% & 30m \\
   \hline
   SA & 112250.94 & 40.26\% & 30m & 248161.94 & 69.82\% & 30m & 416360.91 & 84.94\% & 30m \\
   LNS & 91722.98 & 14.37\% & 30m & 183769.36 & 25.73\% & 30m & 320327.66 & 42.25\% & 30m \\
   LNS\_SA & 94816.34 & 18.45\% & 30m & 183388.98 & 25.44\% & 30m & 324700.14 & 44.16\% & 30m \\
   GA & 143664.25 & 79.45\% & 30m & 304865.51 & 108.75\% & 30m & 515242.92 & 129.10\% & 30m \\
   CP\_LNS & 114576.62 & 43.10\% & 30m & 227609.73 & 55.88\% & 30m & 303815.31 & 34.88\% & 30m \\
   \hline
   Random Insertion & 168649.50 & 96.75\% & 1.70s & 340960.69 & 135.49\% & 0.38m & 612480.69 & 171.80\% & 2.44m \\
   Nearest Insertion & 156261.83 & 82.53\% & 2.33s & 322091.38 & 122.49\% & 0.48m & 590135.81 & 161.90\% & 3.24m \\
   Farthest Insertion & 157519.63 & 83.97\% & 2.65s & 351715.00 & 143.04\% & 0.49m & 666756.19 & 195.92\% & 4.05m \\
   Nearest Neighbor & 147099.97 & 71.44\% & 0.11s & 271017.28 & 87.04\% & 0.25s & 425558.38 & 88.75\% & 0.51s \\
   CWS & 106356.66 & 23.96\% & 0.10s & 190049.50 & 31.08\% & 0.25s & 300500.69 & 33.29\% & 0.50s \\
   \hline\hline
   Ours(Greedy) & 92121.10 & 7.12\% & 2.88s & \textbf{157540.69} & \textbf{8.58\%} & 3.35s & \textbf{242930.86} & \textbf{7.67\%} & 4.08s \\
   Ours(Sampling) & 86380.93 & 0.45\% & 2.98s & \textbf{145099.80} & \textbf{0.02\%} & 3.68s & \textbf{225617.61} & \textbf{0.00\%} & 5.13s \\ 
  \bottomrule
  \end{tabular}
\end{table*}

\subsection{Ablation Study}
\label{exp_ablation}
We first conduct ablation studies to verify the effectiveness of key designs in our neural method. Specifically, we evaluate different loss functions to optimize the policy network for AGH, and different embedding methods for the temporal inputs, where we primarily take AGH100 as the study case. 

\noindent\textbf{Loss Functions.} While the learnt policy for each sub-problem is shared among different fleets, we concern that loss $\mathcal{L}(\theta|x^f)$ in Eq. (\ref{eq:train}) only optimizing the tour length for each sub-problem may ignore the interplay between sub-problems, i.e., the solutions to the higher precedence level affects the time windows of the succeeding sub-problems. Thus, we also extend the loss function to consider the global solution or solutions to a series of sub-problems. Three candidate loss functions are evaluated as follows,

\begin{enumerate}
    \item $\mathcal{L}_{G}$: We optimize the policy network with solutions of all fleets. The gradient is computed as in Eq. (\ref{eq:loss_g}), where $L(a)$ and $\mathcal{B}(x)$ represent costs of the global solutions attained by the current model and baseline model, respectively. This approach optimizes the multi-fleet vehicle routing in AGH as a whole, 
        \begin{gather}
            \label{eq:loss_g}
            \nabla\mathcal{L}_G(\theta|x) = (L(a) - \mathcal{B}(x)) \nabla \log \prod_{f \in \mathcal{F}} \pi_{\theta}(a^f | x^f).
        \end{gather}
    \item $\mathcal{L}_{M_{G}}$: As an alternative, we define the gradient as a weighted sum of the solution cost of each sub-problem and that of the global solution ($\alpha=0.95$), 
        \begin{align}
            \label{eq:loss_m_g}
            \nabla\mathcal{L}_{M_G} & (\theta|x) = \frac{1}{|\mathcal{F}|} \sum_{f \in \mathcal{F}}[(\alpha(L(a^f) - \mathcal{B}(x^f)) + 
            \notag \\ & (1-\alpha)(L(a) - \mathcal{B}(x)))\nabla \log \pi_{\theta}(a^f | x^f)].
        \end{align}
    \item $\mathcal{L}_{M_{F}}$: We further refine $\mathcal{L}_{M_{G}}$ by only considering the solution costs of sub-problems with lower priorities,
        \begin{align}
            \label{eq:loss_m_f}
            & \nabla\mathcal{L}_{M_F}(\theta|x) = \frac{1}{|\mathcal{F}|} \sum_{f \in \mathcal{F}} [(\alpha(L(a^f) - \mathcal{B}(x^f)) + \notag \\ & (1-\alpha)\sum_{p\prec p'}\sum_{f^{'}\in s(p')} (L(a^{f^{'}}) - \mathcal{B}(x^{f'}))) \nabla \log \pi_{\theta}(a^f | x^f)].
        \end{align}
\end{enumerate}
Note that, same as Eq.~(\ref{eq:train}), the above loss functions are used to update the policy network after the global solution to AGH is constructed. Then we train the policies with the above loss functions and test them on AGH100 instances. The results are gathered in TABLE \ref{exp_table_3}, where we observe that the loss function in Eq.~(\ref{eq:train}) achieves better performance than the above three variants. Meanwhile, we display the validation performance during training in Fig. \ref{fig_ablation}. We observe that the training process with the loss $\mathcal{L}_G$ or $\mathcal{L}_{M_G}$ is difficult to be converged. It might result from the relatively high complexity involving the whole AGH solution for $\mathcal{L}_G$, and the noisy information for $\mathcal{L}_{M_G}$. Here, the noisy information refers to the solution costs from operations that are not related much, since the solution to an operation may only affect the ones with lower priorities.
On the other hand, although $\mathcal{L}_{M_F}$ performs better than $\mathcal{L}_{M_G}$, it is still inferior to $\mathcal{L}(\theta|x^f)$ based on our empirical results. Thus we conclude that the loss $\mathcal{L}(\theta|x^f)$ is sufficiently effective and efficient to train a desirable policy to solve AGH.

\noindent\textbf{Temporal Embeddings.} Since the time windows in each operation are affected by solutions to the operations with a higher priority in AGH, we would like to emphasize the importance of embedding such temporal information, and also investigate the recurrent neural network (e.g., LSTM) for time window embedding. Specifically, the temporal input $a_j^p$ and $b_j^p$ are first processed with a linear projection, and then passed to LSTM accompanied with the hidden and cell states. Those hidden and cell states are initialized with zero vectors and updated as the output of LSTM at the last step, which is regarded as the embedding of temporal feature, and then added to the initial embedding $\textbf{h}_{j}^{(0)}$.
As shown in TABLE \ref{exp_table_3} and Fig. \ref{fig_ablation}, the policy learned without representation of the temporal feature (w/o TW) performs significantly inferior to the one with such representation, both for the training and testing. Also, the results indicate that the recurrent neural network does not bring obvious advantage over ours. It only slightly improves the validation performance at the very beginning of the training. Hence, we stick to the usage of the linear projection in Eq.~(\ref{eq:20}) to embed the temporal input for better parallelism and less computational cost.

\subsection{Comparison Study}
\label{exp_comparison}
We compare our neural method with baselines on AGH20, AGH50 and AGH100 and record the average objective value, primal gap and (needed) runtime. All results are displayed in TABLE~\ref{exp_table_1}. We observe that CPLEX is able to solve small instances well on AGH20 with the given 30 minutes, achieving the smallest objective value and primal gap among all methods. However, its performance saliently degrades on larger instances, which is generally inferior to the other methods on AGH50 and AGH100. The reason is that the computational complexity increases exponentially as the size grows due to its NP-hard nature, and CPLEX based on the exact algorithm becomes less effective in searching optimal or high-quality solutions. On the other hand, our neural method can attain the best results on AGH50 and AGH100, and especially when multiple solutions are sampled, the average primal gap is almost 0. It means our method outperforms all baselines on almost all instances. Compared to meta-heuristics, including the two methods (GA and CP\_LNS) specialized for AGH, even our method with \textit{Greedy} decoding achieves better results than them on all problem sizes (except for LNS on AGH20). Moreover, our method with \textit{Sampling} decoding further improves the results significantly. Note that our method is more efficient than meta-heuristics, since the policy network constructs the solution to an instance once without additional steps for iterative improvements. Additionally, while GA and CP\_LNS resort to domain knowledge and trial-and-error to derive the algorithm, e.g., the design of operators, our method can learn more powerful policies automatically. Meanwhile, it is also clear that the learned construction heuristic by our method is considerably superior to classic insertion heuristics and CWS, which again verifies its favorable capability in solving AGH.

\begin{table}[!t]
  \newcommand{\tabincell}[2]{\begin{tabular}{@{}#1@{}}#2\end{tabular}}
  \renewcommand{\arraystretch}{1.5}
  \renewcommand\tabcolsep{3pt}
  \caption{Generalization study on problem sizes.}
  \label{exp_table_2}
  \centering
  \scriptsize
  \begin{tabular}{l|rrr|rrr}
  \toprule
   \multirow{2}*{Method} & \multicolumn{3}{c|}{AGH200} & \multicolumn{3}{c}{AGH300}\\
    & Obj. & Gap & Time & Obj. & Gap & Time\\
   \hline
   LNS & 686158.86 & 86.18\% & 1h & 1576055.58 & 223.49\% & 1h \\
   LNS\_SA & 685258.40 & 85.88\% & 1h & 1521977.52 & 212.27\% & 1h \\
   CP\_LNS & 471051.22 & 27.32\% & 1h & 576481.92 & 18.80\% & 1h \\
   \hline
   Random Insertion & 1163235.50 & 216.39\% & 0.33h & 1684740.15 & 246.08\% & 1.27h \\
   Nearest Insertion & 1110559.75 & 202.08\% & 0.47h & 1577794.82 & 224.13\% & 1.69h \\
   Farthest Insertion & 1297243.25 & 252.82\% & 0.51h & 1895887.19 & 289.45\% & 1.67h \\
   Nearest Neighbor & 664187.06 & 80.58\% & 0.97s & 858120.69 & 76.23\% & 1.43s \\
   CWS & 491486.56 & 33.60\% & 1.01s & 638389.44 & 31.09\% & 1.52s \\
   \hline\hline
   Ours(Greedy) & \textbf{387361.94} & \textbf{5.28\%} & 6.78s & \textbf{504203.06} & \textbf{3.53\%} & 8.55s \\
   Ours(Sampling) & \textbf{367934.56} & \textbf{0.00\%} & 9.89s & \textbf{487027.66} & \textbf{0.00\%} & 17.49s \\ 
  \bottomrule
  \end{tabular}
\end{table}

\subsection{Generalization Study}
\label{exp_generalization}
After the policy is trained, we expect that it can generalize well to larger unseen instances. To demonstrate such a capability, we directly apply the policy trained on AGH100 to solve 1000 instances of AGH200 and AGH300, which are generated following almost the same procedure as described in Section \ref{exp_settings}. The only difference is that we set the capacity of vehicles to 60 and 70 for AGH200 and AGH300, respectively. For meta-heuristics, we only report the results of LNS, LNS\_SA and CP\_LNS with the time limit of 1 hour, since we found that they perform better than SA and GA with a similar pattern in Section \ref{exp_comparison}. We also compare with classic construction heuristics, and all results are summarized in TABLE~\ref{exp_table_2}. It is revealed that CP\_LNS and CWS attain the best results among the meta-heuristics and construction heuristics, respectively.  On the other hand, our method with either \textit{Greedy} or \textit{Sampling} decoding strategy can still yield better solutions than all baselines on both AGH200 and AGH300. Moreover, the needed runtime of the construction heuristics is shorter than that of meta-heuristics, exhibiting a desirable efficiency on the large instances. In summary, our method is able to achieve the best solutions with reasonably short runtime when generalizing to larger AGH instances.

On the other hand, it is also necessary to evaluate the trained model with instances from varying distributions, so as to verify its potential to be used with diverse modes of instance parameters. Although our instance generation mostly follows the real scenarios in Changi airport, we further manifest the power of the model with varying flight demands and arrival time. Specifically, we sample flight demands from Gaussian (with mean 5 and variance 2.5) and Poisson distributions (with the expected number of events 5), respectively. Similarly, we also sample the arrival time of aircraft by assigning probabilities to each hour, which are derived from Gaussian (with mean 0 and variance 1) and Poisson (with the expected number of events 4) distributions. We directly apply the trained model to the instances and record the results in TABLE~\ref{exp_table_21}. It is clear that our methods with Sampling can achieve the lowest gaps and objective values under either Gaussian or Poisson distribution. Furthermore, the runtime of our method is relatively short compared to the metaheuristics and some of the construction heuristics. Despite the good generalization to the above two distributions, we will apply the proposed method to solve AGH in different airports in the future, so as to evaluate it with other practical parameter distributions.

\begin{table}[!t]
  \newcommand{\tabincell}[2]{\begin{tabular}{@{}#1@{}}#2\end{tabular}}
  \renewcommand{\arraystretch}{1.5}
  \renewcommand\tabcolsep{3pt}
  \caption{Generalization study on varying parameters.}
  \label{exp_table_21}
  \centering
  \scriptsize
  \begin{tabular}{l|rrr|rrr}
  \toprule
   \multirow{2}*{Method} & \multicolumn{3}{c|}{Gaussian} & \multicolumn{3}{c}{Poisson}\\
    & Obj. & Gap & Time & Obj. & Gap & Time\\
   \hline
   LNS & 311851.84 & 37.33\% & 30m & 320533.46 & 40.44\% & 30m \\
   LNS\_SA & 309631.53 & 36.33\% & 30m & 316322.71 & 38.89\% & 30m \\
   CP\_LNS & 306331.68 & 34.93\% & 30m & 312833.22 & 37.75\% & 30m \\
   \hline
   Random Insertion & 607621.00 & 166.88\% & 2.79m & 619340.75 & 172.94\% & 3.45m \\
   Nearest Insertion & 584337.69 & 156.64\% & 3.85m & 593679.06 & 161.70\% & 4.66m \\
   Farthest Insertion & 659150.63 & 189.50\% & 5.13m & 674967.44 & 197.53\% & 5.95m \\
   Nearest Neighbor & 424645.47 & 86.38\% & 0.55s & 424585.13 & 87.00\% & 0.57s \\
   CWS & 304361.41 & 33.58\% & 0.57s & 311102.69 & 36.99\% & 0.52s \\
   \hline\hline
   Ours(Greedy) & \textbf{246531.45} & \textbf{8.14\%} & 5.15s & \textbf{245913.59} & \textbf{8.23\%} & 4.89s \\
   Ours(Sampling) & \textbf{227973.09} & \textbf{0.00\%} & 6.87s & \textbf{227209.89} & \textbf{0.00\%} & 6.84s \\ 
  \bottomrule
  \end{tabular}
\end{table}

\subsection{Real-time AGH}
\label{stochastic_agh}
In reality, we not only concern a good solution to AGH that could be computed days earlier, but also hope the model could handle real-time AGH that is characterized with stochastic arrivals of flights. In other words, when the known flights are being scheduled, subsequent (new) flights will arrive at airport at random time which are not known in advance. 
To evaluate this property, we adapt our method and construction heuristics to the real-time AGH setting (in a re-optimization manner), which mimics the real-world dynamics in an airport. When the new flights come, we fix the scheduling of operations before current timestamp, and continue to schedule the vehicle fleets for \emph{unserved} flights.
Following the same instance generation procedure in Section \ref{exp_settings}, we first generate 500 instances of AGH50, and gradually add new incoming flights (until 100 flights in each instance) during inference whose arrival time are sampled according to the statistics of CHANGI airport. Note that the arrival time of new flight cannot precede the current timestamp, and hence is revealed dynamically during the execution of operations.
From the results in TABLE \ref{exp_table_4}, it is exhibited that our method can deal with well the dynamic arrivals of flights and offer shorter route to dispatch all fleets for operations. In addition, our method runs efficiently which meets the requirements of real-time AGH, which well justified its favorable capability of handling dynamics and randomness in practice.


\begin{table}[!t]
  \newcommand{\tabincell}[2]{\begin{tabular}{@{}#1@{}}#2\end{tabular}}
  \renewcommand{\arraystretch}{1.5}
  \renewcommand\tabcolsep{13pt}
  \caption{Results for real-time AGH}
  \label{exp_table_4}
  \centering
  \footnotesize
  \begin{tabular}{l|rrr}
  \toprule
  \multirow{2}*{Method} & \multicolumn{3}{c}{AGH50$\to$100}\\
    & Obj. & Gap & Time\\
  \hline
  Random Insertion & 558672.81 & 102.72\% & 2.02m \\
  Nearest Insertion & 429569.28 & 55.97\% & 2.98m \\
  Farthest Insertion & 429871.59 & 56.11\% & 2.86m \\
  Nearest Neighbor & 515896.97 & 87.60\% & 3.25s \\
  CWS & 336290.59 & 21.99\% & 2.90s \\
  \hline\hline
  Ours(Greedy) & \textbf{305379.37} & \textbf{10.55\%} & 10.99s \\
  Ours(Sampling) & \textbf{276043.84} & \textbf{0.00\%} & 14.90s \\ 
  \bottomrule
  \end{tabular}
\end{table}

\section{Conclusions and Future Works}
\label{conclusion}
In this paper, we first propose a neural method to solve AGH. We present a construction framework for AGH with multiple (types of) operations, which decomposes the studied problem into VRPs for each fleet and solves them with a construction heuristic following the precedence relation. Particularly, we concretize the construction heuristic with an attention-based policy network and train it by the RL algorithm, which is shared by each fleet (sub-problem). Results show that our method outperforms classic construction heuristics, meta-heuristics and existing specialized methods for AGH, in terms of solution quality and computational efficiency. Moreover, our method generalizes well to instances with larger scales or different parameters and performs favorably on real-time AGH with stochastic flight arrivals, which yields superior results to all baselines. 

In addition, the ablation study provides two pivotal insights in the algorithmic design for solving AGH. First, the time window is an important factor that intertwines sub-problems in AGH. The linear projection is empirically found to be better than LSTM to embed this information into the policy network. Second, optimizing the solutions to sub-problems gains superior performance to the direct optimization of the global solution during training. These findings may serve as a source of inspiration for the use of DRL in addressing other AGH variations and similar practical VRPs.

We would like to note that this work is an early attempt to solve AGH with deep learning, which has substantial practical significance. First, it benefits the algorithmic development for AGH in the data-driven fashion, without much trial-and-error and domain knowledge to design hand-crafted rules or operators. Also, the neural method can deliver superior solutions to classic methods in a fairly short time, which has a potential to promote the efficiency of airport management and the economics of aviation. Furthermore, the proposed method belongs to an appealing application of DRL for solving practical VRPs with complex constraints, in comparison to most existing neural methods for simple and standard VRPs. In the future, we will, 1) apply our method to much larger AGH instances such as the ones with 1000 flights; 2) adapt it to other stochastic settings, e.g., the uncertain service time for operations; 3) involve heterogeneous vehicles in fleets.

\bibliographystyle{IEEEtran}
\bibliography{IEEEabrv, camera_ready}



\begin{IEEEbiography}[{\includegraphics[width=1in,height=1.25in,clip,keepaspectratio]{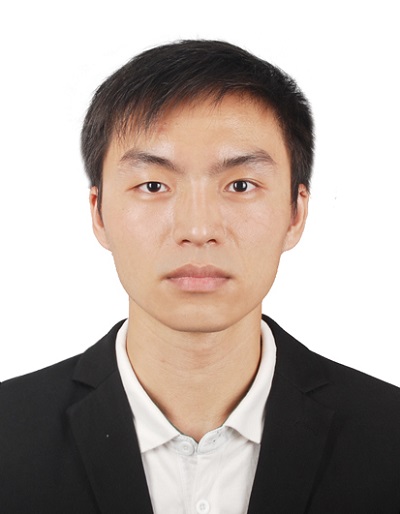}}]{Yaoxin Wu}
received the B.Eng degree in traffic engineering from Wuyi University, Jiangmen, China, in 2015, the M.Eng degree in control engineering from Guangdong University of Technology, Guangzhou, China, in 2018, and the Ph.D. degree in computer science from Nanyang Technological University, Singapore, in 2023. He was a Research Associate with the Singtel Cognitive and Artificial Intelligence Lab for Enterprises (SCALE@NTU). He joins the Department of Information Systems, Faculty of Industrial Engineering and Innovation Sciences, Eindhoven University of Technology, as an Assistant Professor. His research interests include combinatorial optimization, integer programming and deep learning.
\end{IEEEbiography}

\begin{IEEEbiography}[{\includegraphics[width=1in,height=1.25in,clip,keepaspectratio]{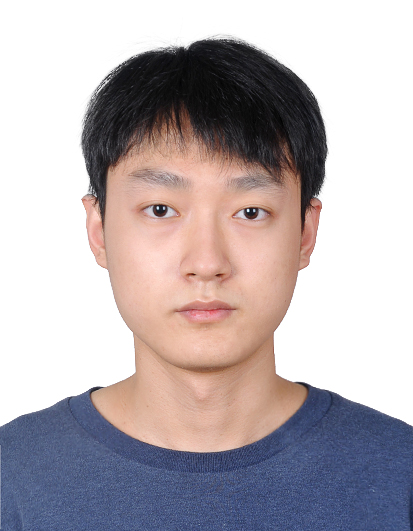}}]{Jianan Zhou}
received the B.Eng. degree in software engineering from Northeastern University, Shenyang, China, in 2019, and the M.Sc. degree in artificial intelligence from Nanyang Technological University, Singapore, in 2021. He is currently pursuing the Ph.D. degree with the School of Computer Science and Engineering, Nanyang Technological University, Singapore. His research interest includes machine learning with combinatorial optimization problems.
\end{IEEEbiography}

\begin{IEEEbiography}[{\includegraphics[width=1in,height=1.25in,clip,keepaspectratio]{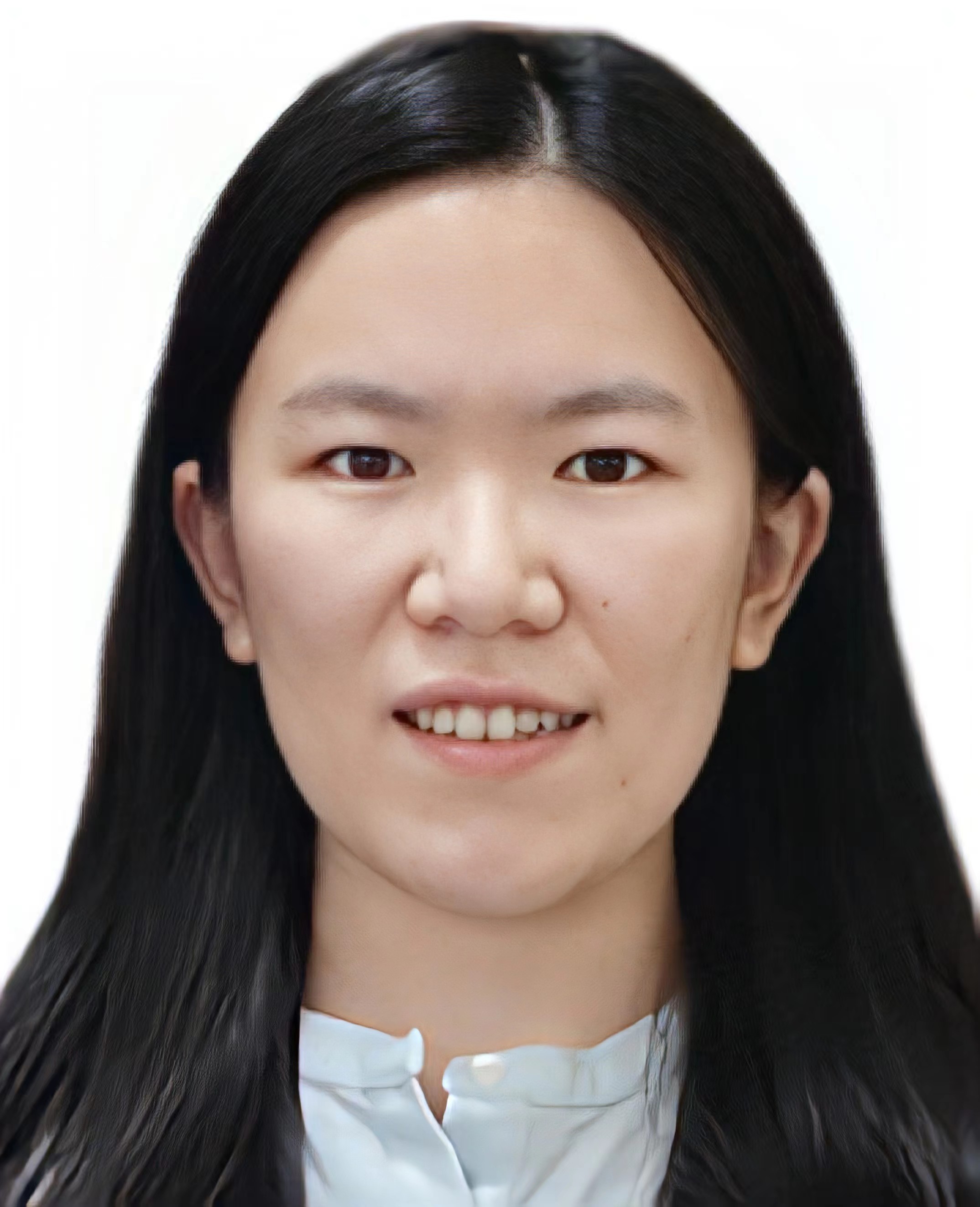}}]{Yunwen Xia}
received the B.Eng. degree in computer science and Technology from Xiamen University, Xiamen, China, in 2019, and the M.Eng degree from Nanyang Technological University, Singapore, in 2022. She was a Project Officier with the Singtel Cognitive and Artificial Intelligence Lab for Enterprises (SCALE@NTU), Nanyang Technological University. Her research interest includes recommendation system and neural combinatorial optimization.
\end{IEEEbiography}

\begin{IEEEbiography}[{\includegraphics[width=1in,height=1.25in,clip,keepaspectratio]{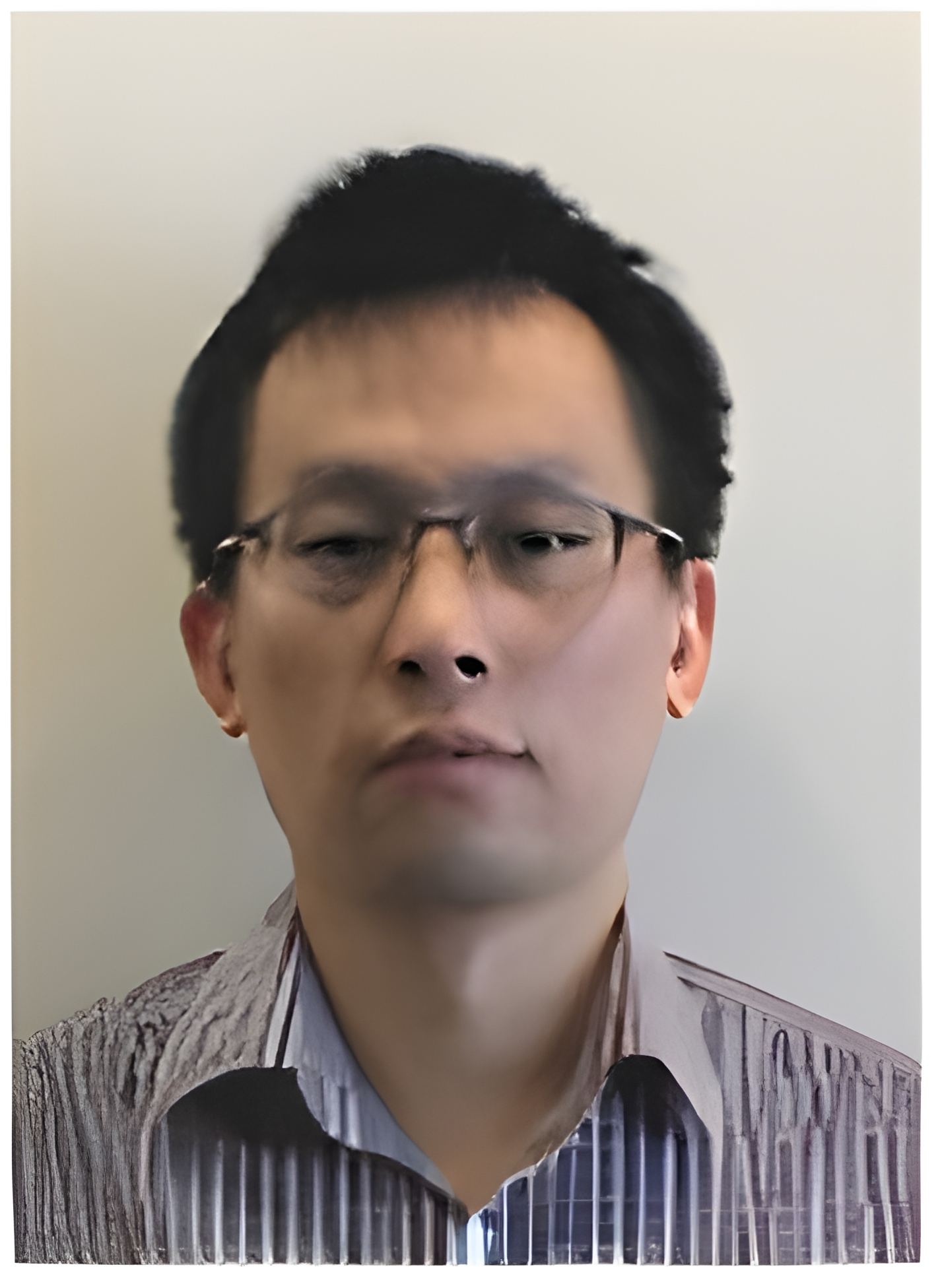}}]{Xianli Zhang}
received the B.Eng. degree in survey engineering from Wuhan University, Wuhan, China, in 2000, and the M.Eng. degree in electronics and communication engineering from Beihang University, Beijing, China, in 2008. He is currently a Ph.D. candidate with the School of Computer Science and Engineering, Nanyang Technological University, Singapore. His research interests include scheduling, evolutionary computation, and deep learning.
\end{IEEEbiography}

\begin{IEEEbiography}[{\includegraphics[width=1in,height=1.25in,clip,keepaspectratio]{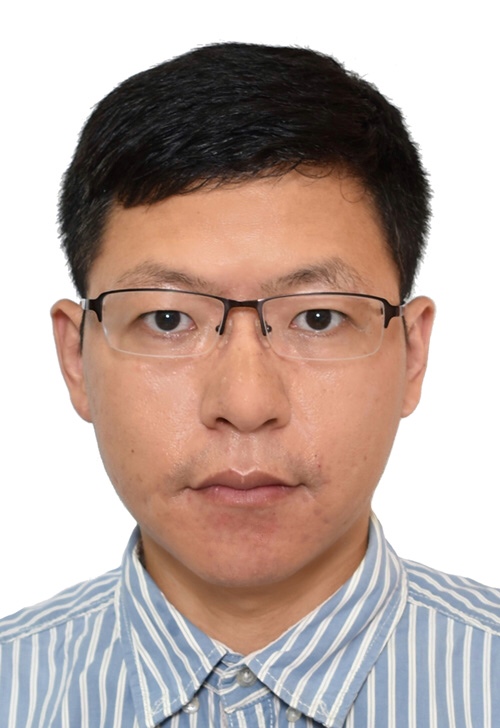}}]{Zhiguang Cao}
received the Ph.D. degree from Interdisciplinary Graduate School, Nanyang Technological University. He received the B.Eng. degree in Automation from Guangdong University of Technology, Guangzhou, China, and the M.Sc. in Signal Processing from Nanyang Technological University, Singapore, respectively. He was a Research Fellow with the Energy Research Institute @ NTU (ERI@N), a Research Assistant Professor with the Department of Industrial Systems Engineering and Management, National University of Singapore, and a Scientist with the Agency for Science Technology and Research (A*STAR), Singapore. He joins the School of Computing and Information Systems, Singapore Management University, as an Assistant Professor. His research interests focus on learning to optimize (L2Opt).
\end{IEEEbiography}

\begin{IEEEbiography}[{\includegraphics[width=1in,height=1.25in,clip,keepaspectratio]{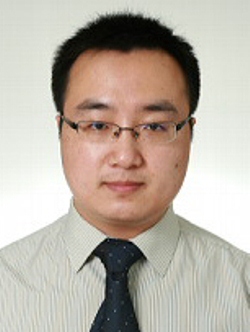}}]{Jie Zhang}
received the Ph.D. degree from the Cheriton School of Computer Science, University of Waterloo, Canada, in 2009. He is currently a Professor with the School of Computer Science and Engineering, Nanyang Technological University, Singapore. He is also a Professor at the Singapore Institute of Manufacturing Technology. During his Ph.D. study, he held the prestigious NSERC Alexander Graham Bell Canada Graduate Scholarship rewarded for top Ph.D. students across Canada. He was also a recipient of the Alumni Gold Medal at the 2009 Convocation Ceremony. The Gold Medal is awarded once a year to honour the top Ph.D. graduate from the University of Waterloo. His papers have been published by top journals and conferences and received several best paper awards.
\end{IEEEbiography}

\end{document}